%% file: main.tex
\documentclass[runningheads]{llncs}
\usepackage{graphicx}

\usepackage{tikz}
\usetikzlibrary{arrows,decorations.markings,shapes,arrows,fit}
\usetikzlibrary{positioning}
\usepackage[subpreambles=true]{standalone}

\usepackage{amsmath,amssymb}
\usepackage{color}

\usepackage[accsupp]{axessibility}

\usepackage{graphicx}
\usepackage{multirow}
\usepackage{booktabs}
\usepackage{graphics}
\usepackage[colorlinks=true,citecolor=blue,urlcolor=black]{hyperref}
\usepackage{enumitem}
\usepackage{makecell}
\usepackage{url}
\usepackage[font=footnotesize]{caption}
\usepackage{subcaption}
\usepackage{appendix}
\usepackage{tabularx}

\makeatletter
\def\@seccntformat#1{\@ifundefined{#1@cntformat}
   {\csname the#1\endcsname\quad}
   {\csname #1@cntformat\endcsname}
}
\let\oldappendix\appendix
\renewcommand\appendix{
    \oldappendix
    \newcommand{\section@cntformat}{\appendixname~\thesection\quad}
}
\makeatother
\usepackage[capitalise,nameinlink]{cleveref}

\usepackage{tikz}
\usetikzlibrary{bayesnet, calc, arrows.meta, bending}
\usetikzlibrary{decorations.pathreplacing, angles, quotes, shapes.multipart, calligraphy}

\tikzset{x=13mm, y=13mm, on grid=true, >=Latex}
\tikzstyle{latent} += [minimum size=16pt, inner sep=0pt]
\tikzstyle{invertible} = [{Latex[open]}-Latex]
\tikzstyle{cvar} = [draw, fill, circle, inner sep=0pt, minimum size=3pt]
\tikzstyle{obsdet} = [det, fill=gray!25]
\tikzstyle{amort} = [densely dotted, >={Latex[open]}]
\tikzstyle{aux} = [draw, fill=white, circle, minimum size=2pt, inner sep=0pt]

\newcommand{\markerone}{\raisebox{0.5pt}{\tikz{\node[draw,scale=0.6,regular polygon, regular polygon sides=4,fill=black!20!blue](){};}}}

\newcommand{\markertwo}{\raisebox{0.5pt}{\tikz{\node[draw,scale=0.6,regular polygon, regular polygon sides=4,fill=black!10!red](){};}}}

\newcommand{\mech}{f}

\newcommand{\noisehi}{$z_X$}
\newcommand{\noiselox}{$u_X$}
\newcommand{\noisehix}{$z_X$}

\newcommand{\chebblockdown}{\operatorname{ChebBlock}^\downarrow_{\theta}}
\newcommand{\chebconv}{\operatorname{ChebConv}_{\theta}}

\newcommand{\condenc}{\operatorname{CondEnc}_{X}}
\newcommand{\conddec}{\operatorname{CondDec}_{X}}
\newcommand{\quadricup}{\operatorname{Simplify}\inv}
\newcommand{\quadricdown}{\operatorname{Simplify}}
\newcommand{\elu}{\operatorname{ELU}}

\newcommand{\cf}{\mathrm{cf}}

\newcommand{\defeq}{:=}
\newcommand{\inv}{^{-1}}

\begin{document}

\pagestyle{headings}
\mainmatter

\title{Deep Structural Causal Shape Models}

\titlerunning{Deep Structural Causal Shape Models}

\author{Rajat Rasal\inst{1} \and
Daniel C. Castro\inst{2}\index{Castro, Daniel C.} \and
Nick Pawlowski\inst{2} \and
Ben Glocker\inst{1}}

\institute{Department of Computing, Imperial College London, UK \and Microsoft Research, Cambridge, UK}
\maketitle

\begin{abstract}
Causal reasoning provides a language to ask important interventional and counterfactual questions beyond purely statistical association. In medical imaging, for example, we may want to study the causal effect of genetic, environmental, or lifestyle factors on the normal and pathological variation of anatomical phenotypes. However, while anatomical shape models of 3D surface meshes, extracted from automated image segmentation, can be reliably constructed, there is a lack of computational tooling to enable causal reasoning about morphological variations. To tackle this problem, we propose deep structural causal shape models (CSMs), which utilise high-quality mesh generation techniques, from geometric deep learning, within the expressive framework of deep structural causal models. CSMs enable subject-specific prognoses through counterfactual mesh generation (``How would this patient's brain structure change if they were ten years older?''), which is in contrast to most current works on purely population-level statistical shape modelling. We demonstrate the capabilities of CSMs at all levels of Pearl's causal hierarchy through a number of qualitative and quantitative experiments leveraging a large dataset of 3D brain structures.

\end{abstract}

\section{Introduction}
The causal modelling of non-Euclidean structures is a problem which machine learning research has yet to tackle. Thus far, state-of-the-art causal structure learning frameworks, utilising deep learning components, have been employed to model the data generation process of 2D images \cite{pawlowski2020dscm,kocaoglu2017causalgan,sauer2021counterfactual,yang2021causalvae}. However, most of these approaches fall short of answering counterfactual questions on observed data, and to the best of our knowledge, none have been applied to non-Euclidean data such as 3D surface meshes. Parallel streams of research have rapidly advanced the field of geometric deep learning, producing highly performant predictive and generative models of graphs and meshes \cite{bronstein2017geometric,bronstein2021geometric,egger20203d,guo2020systematic,ZHOU202057,wu2020comprehensive}. Our work finds itself at the intersection of these cutting-edge fields; our overarching contribution is a deep structural causal shape model (CSM), utilising geometric deep learning components, of the data generation process of 3D meshes, with tractable counterfactual mesh generation capabilities.

Traditional statistical learning techniques only allow us to answer questions that are inherently associative in nature. For example, in a supervised learning setting, we are presented with a dataset of independently and identically distributed features, from which we pick some to be the inputs $x$, in order to learn a mapping to targets $y$ using an appropriate machine learning model, such as a neural network. It need not be the case however, that $x$ \textit{caused} $y$ for their statistical relationship to be learned to high levels of predictive accuracy. In order to answer many \textit{real-world} questions, which are often interventional or counterfactual, we must consider the direction of causality between the variables in our data. This is particularly important in the context of medical imaging \cite{castro2020causality}, for example, where one may need to compare the effect of a disease on the entire population against the effect on an individual with characteristic genes, or other important influencing factors, with all assumptions available for scrutiny.

Pearl's causal hierarchy \cite{pearl2019seven,glymour2016causal} is a framework for categorising statistical models by the questions that they can answer. At the first level in the hierarchy, associational models learn statistical relationships between observed variables, e.g.\ $p(y|x)$. In the case of deep generative modelling, for example, variational autoencoders (VAE) \cite{kingma2013auto}, normalising flows (NF) \cite{rezende2015variational,papamakarios2021normalizing} and generative adversarial networks (GAN) \cite{goodfellow2014generative} all learn correlations between a latent variable and the high-dimensional variable of interest, either implicitly or explicitly. Similarly, in geometric deep learning, 3D morphable models \cite{egger20203d,ranjan2018generating,bouritsas2019neural,gong2019spiralnet,tretschk2020demea,zhou2020fully,chen2021learning,Booth_2016_CVPR,cheng2019meshgan} associate a latent vector to a 3D shape. At the next level, interventions can be performed on the assumed data generating process, to simulate an event at the population-level, by fixing the output of a variable generating function \cite{richardson2013single}. In the simplest case, this amounts to interpolating an independent, latent dimension in a statistical shape model \cite{kim2018disentangling,higgins2016beta,chen2018isolating}, with prior work to semantically interpret each latent dimension. Structural causal models (SCMs) (\cref{sec:dscm_mechanisms}) can also be built using neural networks to enable this functionality \cite{yang2021causalvae,kocaoglu2017causalgan,sauer2021counterfactual}. At the final level, counterfactual questions can be answered by using SCMs to simulate subject-specific, retrospective, hypothetical scenarios \cite{glymour2016causal}. Although some work has been done in this field \cite{kocaoglu2017causalgan,sauer2021counterfactual}, Pawlowski et al.'s deep structural causal models (DSCM) framework \cite{pawlowski2020dscm} is the only one to provide a full recipe for tractable counterfactual inference for high and low-dimensional variables, to the best of our knowledge. However, they only considered 2D Euclidean data (images).

\paragraph{\bf Contributions.} The key contributions of this paper are: \textbf{1)} the introduction of the first causally grounded 3D shape model (\cref{sec:dscm_architecture}), utilising advanced geometric deep learning components, capable of inference at all levels of Pearl's hierarchy, including counterfactual queries;
\textbf{2)} a mesh conditional variational autoencoder applied to a biomedical imaging application of subcortical surface mesh generation (\cref{sec:coma}) producing high-quality, anatomically plausible reconstructions (\cref{sec:associational_results});
\textbf{3)} quantitative and qualitative experiments demonstrating the capabilities of the proposed CSM for both population-level (\cref{sec:interventional_results}) and, importantly, subject-specific (\cref{sec:individual_specific_shapes}) causal inference. We demonstrate that CSMs can be used for answering questions such as ``How would \emph{this} patient's brain structure change if they had been ten years older or from the opposite (biological) sex?'' (\cref{fig:counterfactual_interpolation_main}). Our work may be an important step towards the goal of enabling subject-specific prognoses and the simulation of population-level randomised controlled trials.

\section{Related Work}
\label{sec:related_work}

\paragraph{\bf Graph Convolutional Networks.} 
Bruna et al. \cite{bruna2013spectral} proposed the first CNN-style architecture using a spectral convolutional operator \cite{chung1997spectral}, which required the eigendecomposition of the input graph's Laplacian. Defferrard et al. \cite{defferrard2016convolutional} used $k$-truncated Chebyshev polynomials to approximate the graph Laplacian, preventing an eigendecomposition, resulting in a $k$-localised, efficient, linear-time filter computation. Spatial CNNs \cite{monti2017geometric,boscaini2016learning,fey2018splinecnn} outperform spectral CNNs on mesh and node classification tasks, and are domain-independent. We utilise fast spectral convolutions, following their success in deep 3D shape modelling \cite{ranjan2018generating,cheng2019meshgan}.

\paragraph{\bf Encoder-Decoder Frameworks for 3D Shape Modelling.}
Cootes et al. \cite{cootes1995active} introduced PCA-based shape models leading to early applications such as facial mesh synthesis \cite{blanz99morphable}. Booth et al. \cite{Booth_2016_CVPR,booth2018large} presented a ``large-scale" 3D morphable model learned from 10,000 facial meshes. Our dataset is similar in size with a diverse set of brain structures. We use deep learning components for additional expressivity. Litany et al. \cite{litany2018deformable} proposed the first end-to-end mesh autoencoder using graph convolutions from \cite{verma2018feastnet}. Ranjan et al. \cite{ranjan2018generating} then introduced the concept of vertex \textit{pooling} using quadric matrices and switched to the Chebyshev polynomial based operator \cite{defferrard2016convolutional}. We learn localised, multi-scale representations, similar to \cite{ranjan2018generating}, unlike the global, PCA-based methods. The generality of our approach allows for state-of-the-art mesh VAE architectures \cite{bouritsas2019neural,gong2019spiralnet,tretschk2020demea,zhou2020fully,chen2021learning,hahner2022mesh} to be used as drop-in replacements. Barring disparate works \cite{ma2020learning,tan2018variational}, there is little research on conditional mesh VAEs \cite{sohn2015learning}. 3D shape models are often presented on applications of facial \cite{ranjan2018generating,cheng2019meshgan}, hand \cite{mano2017embodied,kulon2019single}, and full body \cite{smpl2015loper,Bogo:CVPR:2014,dfaust:CVPR:2017} datasets.

\paragraph{\bf Causal Deep Learning for Generative Modelling.}
Deep learning has been used for end-to-end causal generation of 2D images from observational data. Kocaoglu et al. \cite{kocaoglu2017causalgan} implemented an SCM by using deep generative networks to model functional mechanisms. Sauer and Geiger \cite{sauer2021counterfactual} improved upon this by jointly optimising the losses of all high-dimensional mechanisms. Parafita and Vitri \cite{parafita2019explaining} model \textit{images} of 3D shapes, but not in an end-to-end fashion. These works, however, were unable to generate counterfactuals from observations, since a framework for abduction was not defined. Pawlowski et al. \cite{pawlowski2020dscm} proposed the deep structural causal model framework (DSCMs) to perform inference at all levels of the causal hierarchy \cite{pearl2019seven}. Causal modelling of non-Euclidean data has not yet been explored, however. DSCMs can include implicit likelihood mechanisms \cite{dash2020counterfactual}, although we opt for explicit mechanisms for ease of training.

\paragraph{\bf Counterfactuals in Medical Image Computing.}
By incorporating generative models into DSCMs, Reinhold et al. \cite{reinhold2021structural} visualised subject-specific prognoses of lesions by counterfactual inference. Similarly, \cite{pawlowski2020dscm} and \cite{mouches2021unifying} used DSCMs to demonstrate subject-specific, longitudinal shape regression of 2D brain images, using observational data. Our work applies the DSCM framework to model 3D meshes. Many recent works on spatiotemporal neuroimaging require longitudinal data \cite{ravi2019degenerative,bone2018learning}, model only population-level, average shapes \cite{huizinga2018spatio}, or do not consider causality in the imaged outcome \cite{xia2021learning}.

\section{Deep Structural Causal Shape Models}
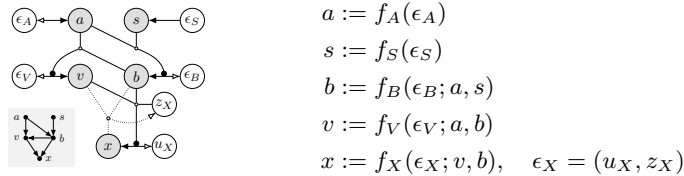
\begin{figure}[b]
    \centering
    \begin{minipage}[c]{.45\textwidth}
    \centering
        \resizebox{0.5\linewidth}{!}{\input{tikz/sem}}
    \end{minipage}
    \begin{minipage}[c]{.3\textwidth}
    \centering
    \vspace{-\baselineskip}
    \begin{align*}
        a &:= f_A(\epsilon_A) \\
        s &:= f_S(\epsilon_S) \\
        b &:= f_B(\epsilon_B; a, s) \\
        v &:= f_V(\epsilon_V; a, b) \\
        x &:= f_X(\epsilon_X; v, b), \quad \epsilon_X = (u_X, z_X)
    \end{align*}
    \end{minipage}
    \caption{Computational graph (\textbf{left}), graphical model (\textbf{bottom-left}) and structural causal model (\textbf{right}). Variables are brain stem mesh~($x$), age~($a$), sex~($s$), and total brain~($b$) and brain stem~($v$) volumes. Reproduced with permission from \cite{pawlowski2020dscm}. Refer to Figure 1 in \cite{pawlowski2020dscm} for a key for the computational graph.}
    \label{fig:graph_ukbb_covariate_sem}
\end{figure}

Our deep structural causal shape model builds upon the DSCM framework \cite{pawlowski2020dscm}. We contextualise this work using a real-world biomedical imaging application leveraging a large dataset of 3D brain structures; we model the shape of a triangulated surface mesh of a person's brain stem ($x$), given their age ($a$), biological sex ($s$), total brain volume ($b$), and brain stem volume ($v$), using observational data from the UK Biobank Imaging Study \cite{sudlow2015uk} (see Appendix for details). We assume the causal graph given in \cref{fig:graph_ukbb_covariate_sem}. Our model is capable of subject-specific and population-level 3D mesh generation as a result of a causal inference process.

\subsection{Counterfactual Inference of High-Dimensional Variables}
\label{sec:dscm_mechanisms}
\paragraph{\bf Structural Causal Models (SCM).} An SCM is defined by the tuple $\\ (\mathcal{E}, p(\mathcal{E}), X, F)$ where $X = \{x_1, \dots, x_M\}$ are covariates of interest, $\mathcal{E} = \{\epsilon_1, \dots, \epsilon_M\}$ are the corresponding exogenous variables with a joint distribution $p(\mathcal{E}) = \prod_{m = 1}^{M} p(\epsilon_m)$, and $F = \{f_1, \dots, f_M\}$ are functional mechanisms which model causal relationships as assignments $x_m := f_m(\epsilon_m; pa_m)$, where $pa_m \subseteq X \setminus \{x_m\}$ are the direct causes (parents) of $x_m$. The stochasticity in generating $x_m$ using $f_m(\cdot)$ is equivalent to sampling $p(x_m | pa_m)$, as per the causal Markov condition, such that $p(X) = \prod_{m = 1}^{M} p(x_m | pa_m)$ \cite{peters2017elements}. Functional mechanisms can therefore be represented in a Bayesian network using a directed acyclic graph. An intervention on $x_m$, denoted $do(x_m := b)$, fixes the output of $f_m(\cdot)$ to $b$ \cite{richardson2013single}.

\paragraph{\bf Deep Structural Causal Models (DSCM).} The DSCM framework \cite{pawlowski2020dscm} provides formulations for functional mechanisms, utilising deep learning components, with tractable counterfactual inference capabilities. High-dimensional data can be generated using amortised, explicit mechanisms as $x_m := f_m(\epsilon_m; pa_m)\\ = l_m(u_m; h_m(z_m; pa_m), pa_m)$, where $\epsilon_m = (u_m, z_m)$, $l_m(\cdot)$ is invertible and $h_m(\cdot)$ is non-invertible. Since DSCMs assume unconfoundedness, the counterfactual $x_{m,\cf}$ can be generated from observations ($x_m$, $pa_m$) as follows:
\begin{enumerate}
    \item Abduction: Infer $p(\epsilon_m|x_m, pa_m)$ by sampling $z_m$ from the variational posterior $q(z_m |\; x_m, pa_m)$ and then calculating $u_m = l_m\inv(x_m; h_m(z_m; pa_m), pa_m)$ exactly. This roughly corresponds to $\epsilon_m = f_m\inv(x_m; pa_m)$.
    \item Action: Perform interventions, e.g.\ $do(x_j := b)$, such that $pa_m$ becomes $\hat{pa}_m$.
    \item Prediction: Sample the counterfactual distribution using the functional mechanism as $x_{m, \cf} = f_m(\epsilon_m; \hat{pa}_m)$.
\end{enumerate}

\subsection{Implementing Causal Shape Models}
\label{sec:dscm_architecture}

\paragraph{\bf Mesh Mechanism.} When applying the DSCM framework to model the causal generative process from \cref{fig:graph_ukbb_covariate_sem} for the mesh $x \in \mathbb{R}^{|V| \times 3}$, $x \sim {p(x | b, v)}$, we define the amortised, explicit mechanism 
\begin{gather}
    x := f_X(\epsilon_X; v, b) = l_X(u_X; \conddec(z_X; v, b)),
    \label{eq:generating_x}
\end{gather}
where $h_X(\cdot) = \conddec(\cdot)$ employs a sequence of spectral graph convolutions with upsampling to predict the parameters of a Gaussian distribution, and $l_X(\cdot)$ performs a linear reparametrisation. Here, there are $|V|$ vertices in each mesh, $z_X \sim \mathcal{N}(0, I_D)$ encodes the mesh's shape, and $u_X \sim \mathcal{N}(0, I_{3|V|})$ captures its scale. Implementation details can be found in \cref{fig:architecture}.

To find $z_X$ during abduction, we encode the mesh $x$ conditioning on its causal parents $(v, b)$, 
\begin{align}
    &(\mu_Z, \log(\sigma_Z^2)) = \condenc(x; v, b), \label{eq:cond_mesh_encoding}
\end{align}
where $\condenc(\cdot)$ uses a sequence of spectral graph convolutions
with downsampling to parametrise an independent Gaussian variational posterior ${q(z_X |\; x, b, v)}$ = ${\mathcal{N}(\mu_Z, \mathrm{diag}(\sigma_Z^2))}$. Then, $u_X$ is computed by inverting the reparametrisation.

$f_V(\cdot)$ and $f_B(\cdot)$ are implemented with the conditional normalising flow
\begin{align}
    \star = \big(\text{Normalisation} \circ \text{ConditionalAffine}_{\theta}(pa_{\star}) \big)(\epsilon_{\star}),
\end{align}
where $\hat{\star} = \text{Normalisation}\inv(\star)$. $\conddec(\cdot)$ and $\condenc(\cdot)$ both utilise intermediate representations $\hat{v}$ and $\hat{b}$ for conditioning, instead of $v$ and $b$, to stabilise training. The conditional location and scale parameters in $\mathrm{ConditionalAffine}_{\theta}(\cdot)$ are learned by shallow neural networks.

\paragraph{\bf Objective.} Mechanisms in our CSM are learned by maximising the joint log-evidence $\log p(x, b, v, a, s)$ at the associational level in the causal hierarchy. From the causal graph in \cref{fig:graph_ukbb_covariate_sem}, ${p(x, b, v, a, s)} = {p(x | b, v)} \cdot {p(v | a, b)} \cdot {p(b | s, a)} \cdot {p(a)} \cdot {p(s)} = {p(x | b, v)} \cdot {p(v,b,a,s)}$. After taking logs, the log-evidence can be written as
\begin{align}
    &\log p(x, b, v, a, s) = \alpha + \log p(x | b, v) = \alpha + \log \int p(x | z_X, b, v) \cdot p(z_X) \; dz_X,
\end{align}
where $\alpha = \log p(v,b,a,s)$. Since the marginalisation over $z_X$ is intractable, we introduce the variational posterior $q(z_X|x,v,b)$, and then formulate an evidence lower bound, by applying Jensen's inequality, that can be optimised,
\begin{align}
    & \log p(x, b, v, a, s)  \label{eq:elbo_1} \; \\
    &\quad \geq \alpha + \underbrace{{E_{q(z_X|x,b,v)}[\log p(x | z_X, b, v)] - \mathrm{KL}[q(z_X | x, b, v) \| p(z_X)]}}_{\beta}, \notag
\end{align}
where ${p(z_X)=\mathcal{N}(0, I_{D})}$ and $\beta$ learns a mesh CVAE \cite{sohn2015learning}, utilising $\conddec(\cdot)$ and $\condenc(\cdot)$, within the CSM structure.

\subsection{Convolutional Mesh Conditional Variational Autoencoder}
\label{sec:coma}

\paragraph{\bf Fast Spectral Convolutions.} Defferrard et al. \cite{defferrard2016convolutional} proposed using Chebyshev polynomials to implement spectral graph convolutions. A kernel $\tau(\cdot; \theta_{j,i})$ is convolved with the $j^\text{th}$ feature in $x \in \mathbb{R}^{|V| \times F_1}$ to produce the $i^\text{th}$ feature in $y \in \mathbb{R}^{|V| \times F_2}$,
\begin{align}
    y_i = \sum^{F_1}_{j=1} \tau(L; \theta_{j,i})x_j = \sum^{F_1}_{j=1} \sum^{K-1}_{k=0} (\theta_{j,i})_k T_k(\tilde{L}) x_j,
\end{align}
where $T_k(\cdot)$ is the $k^\text{th}$ order Chebyshev polynomial, $L$ is the normalised graph Laplacian, $\tilde{L} = 2L\lambda_{|V|}\inv - I_{|V|}$ and $\lambda_{|V|}$ is the largest eigenvalue of $L$. Our meshes are sparse, so $y$ is computed efficiently in $\mathcal{O}(K |V| F_1 F_2)$. For a 3D mesh, $F_1 = 3$ since vertex features are the Cartesian coordinates. We denote this operator as $\chebconv(F, K)$, where $F = F_2$ are the number of output features, and the $K$-truncated polynomial localises the operator to $K$-hops around each vertex.

\paragraph{\bf Mesh Simplification.} The surface error around a vertex $v$ is defined as $v^TQ_vv$, where $Q_v$ is an error quadric matrix \cite{garland1997surface}. For $(v_a, v_b) \rightarrow \hat{v}$ to be a contraction, we choose the $\hat{v} \in \{v_a, v_b, (v_a + v_b) / 2\}$ that minimises the surface error $e = {\hat{v}^T(Q_a + Q_b)\hat{v}}$ by constructing a min-heap of tuples $(e, \hat{v}, v_a, v_b)$. Once the contraction is chosen, adjacent edges and faces are collapsed \cite{ranjan2018generating,ma2020learning,yuan2020mesh,cheng2019meshgan}. We denote the down-sampling operation as $\quadricdown(S)$, with $S > 1$, where $N$ and $\lceil N \cdot \frac{1}{S} \rceil$ are the numbers of vertices before and after simplification, respectively. To perform the corresponding up-sampling operation, $\quadricup(S)$, the contractions are reversed, $\hat{v} \rightarrow (v_a, v_b)$.

\begin{figure}[tb!]
    \centering
    \includegraphics[width=.6\textwidth]{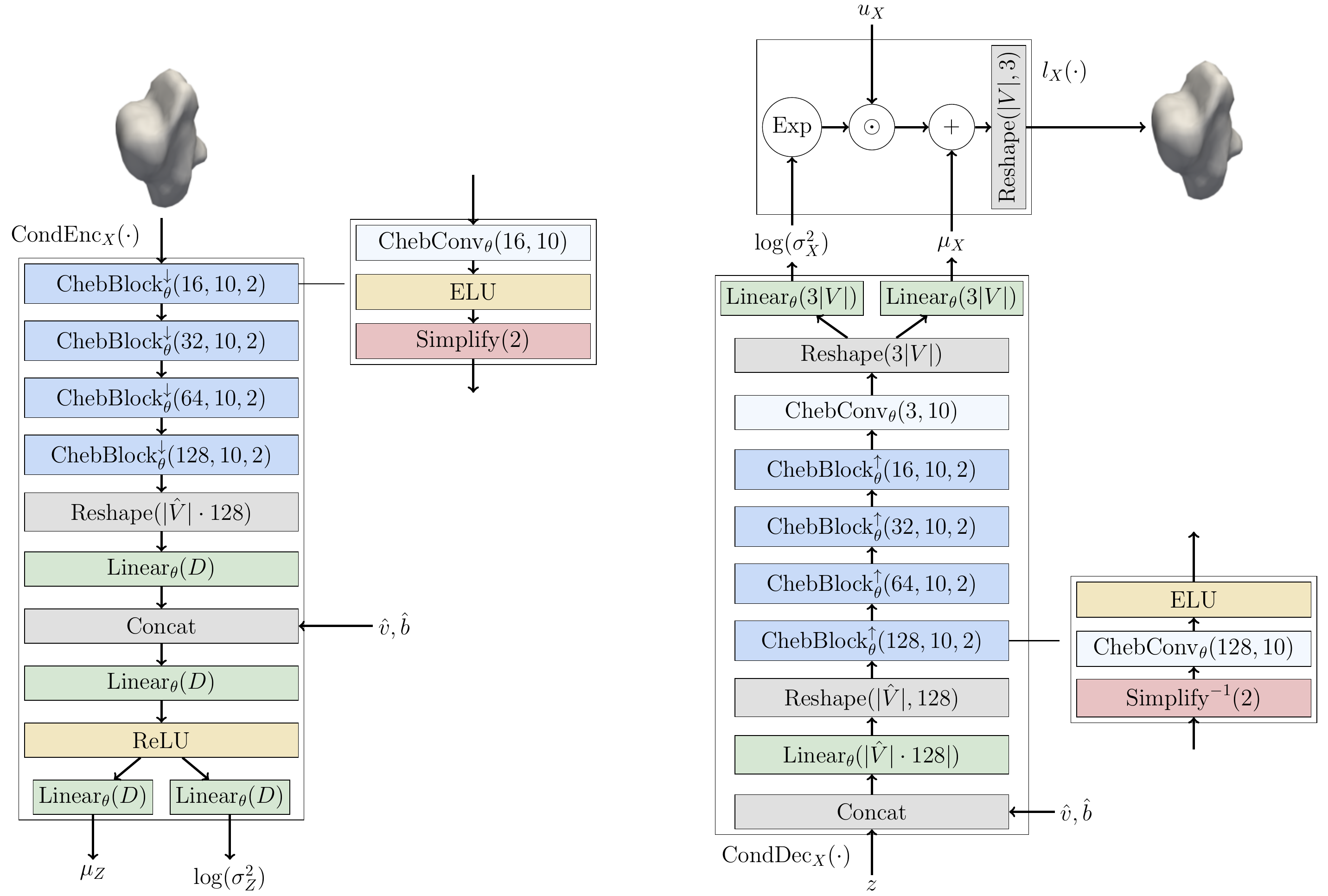}
    \caption{Network architectures for $\condenc(\cdot)$, $h_X(\cdot) = \conddec(\cdot)$, and $l_X(\cdot)$ that are used for $f_X(\cdot)$. $\text{Reshape}(t)$ reshapes the input to the shape given by tuple $t$. $\text{Linear}_{\theta}(M)$ is a fully connected layer with $M$ output features. $\text{ReLU}(\cdot)$ refers to a rectified linear unit \cite{nair2010rectified}. $\text{ELU}(\cdot)$ refers to an exponential linear unit \cite{clevert2015fast}. $|\hat{V}|$ are the number of vertices output after the $\chebblockdown(128,10,2)$ in $\condenc(\cdot)$.}
    \label{fig:architecture}
\end{figure}

\paragraph{\bf Architecture (\cref{fig:architecture}).} 
Brain stem meshes from the UK Biobank are more regular, have far fewer vertices and have a great deal of intraspecific variation when compared to the facial meshes in \cite{ranjan2018generating,cheng2019meshgan}. As a result, we increase the value of $K$ in each $\chebconv(\cdot)$ from 6 to 10, which improves robustness to subtle, localised variations around each vertex. In isolation this can smooth over subject-specific features, especially given that our meshes have a much lower resolution than the aforementioned face meshes. We overcome this by reducing the pooling factor $S$ from 4 to 2. The encoder and decoder are more heavily parametrised and include an additional $\chebconv(3, 10)$ in the decoder, to improve reconstructions from a heavily conditioned latent space. We also utilise $\elu(\cdot)$ activation functions \cite{clevert2015fast} within each $\text{ChebBlock}_{\theta}(\cdot)$, instead of biased ReLUs as in \cite{ranjan2018generating}, to speed up convergence and further improve reconstructions.

\section{Experiments}

\label{sec:experiments}
We analyse meshes generated by the CSM at all levels of the causal hierarchy. Note that the medical validity of the results is conditional on the correctness of the assumed SCM in \cref{fig:graph_ukbb_covariate_sem}, however this should not detract from the computational capabilities of our model. Unless stated otherwise, diagrams are produced using $D = 32$ using data from the test set. Refer to the supplementary material for verification that our model has faithfully learned the data distributions, with further details about the experimental setup, and additional results.


\subsection{Setup}
\paragraph{\bf Preprocessing.}
Each mesh is registered to a template mesh by using the Kabsch--Umeyama algorithm \cite{kabsch1976solution,umeyama1991least} to align their vertices. This removes global effects from image acquisition and patient positioning, leaving primarily shape variations of biological interest while reducing the degrees of freedom $f_X(\cdot)$ has to learn.

\paragraph{\bf Training.}
Our framework is implemented using the Pyro probabilistic programming language \cite{bingham2018pyro}, and the optimisation in \cref{eq:elbo_1} is performed by stochastic variation inference (SVI), using Adam \cite{kingma2014adam}. The inherent noisiness of SVI is magnified when learning high-dimensional, diverse mesh shapes, especially when utilising a Gaussian decoder (L2) over the Laplacian decoder (L1) assumed in \cite{ranjan2018generating}. Using large batch sizes of 256 significantly improves training stability. Mechanisms learn errors at different scales by employing a larger learning rate of $10^{-3}$ for covariate mechanisms compared to $10^{-4}$ for $f_X(\cdot)$. Only 1 Monte-Carlo particle is used in SVI to prevent large errors from $f_X(\cdot)$ causing overflow in the early stages of training. The CSM is trained for 1000 epochs, or until convergence.


\subsection{Associational Level: Shape Reconstruction}
\label{sec:associational_results}

\begin{figure}[t]
    \begin{center}
        \includegraphics[width=0.6\textwidth]{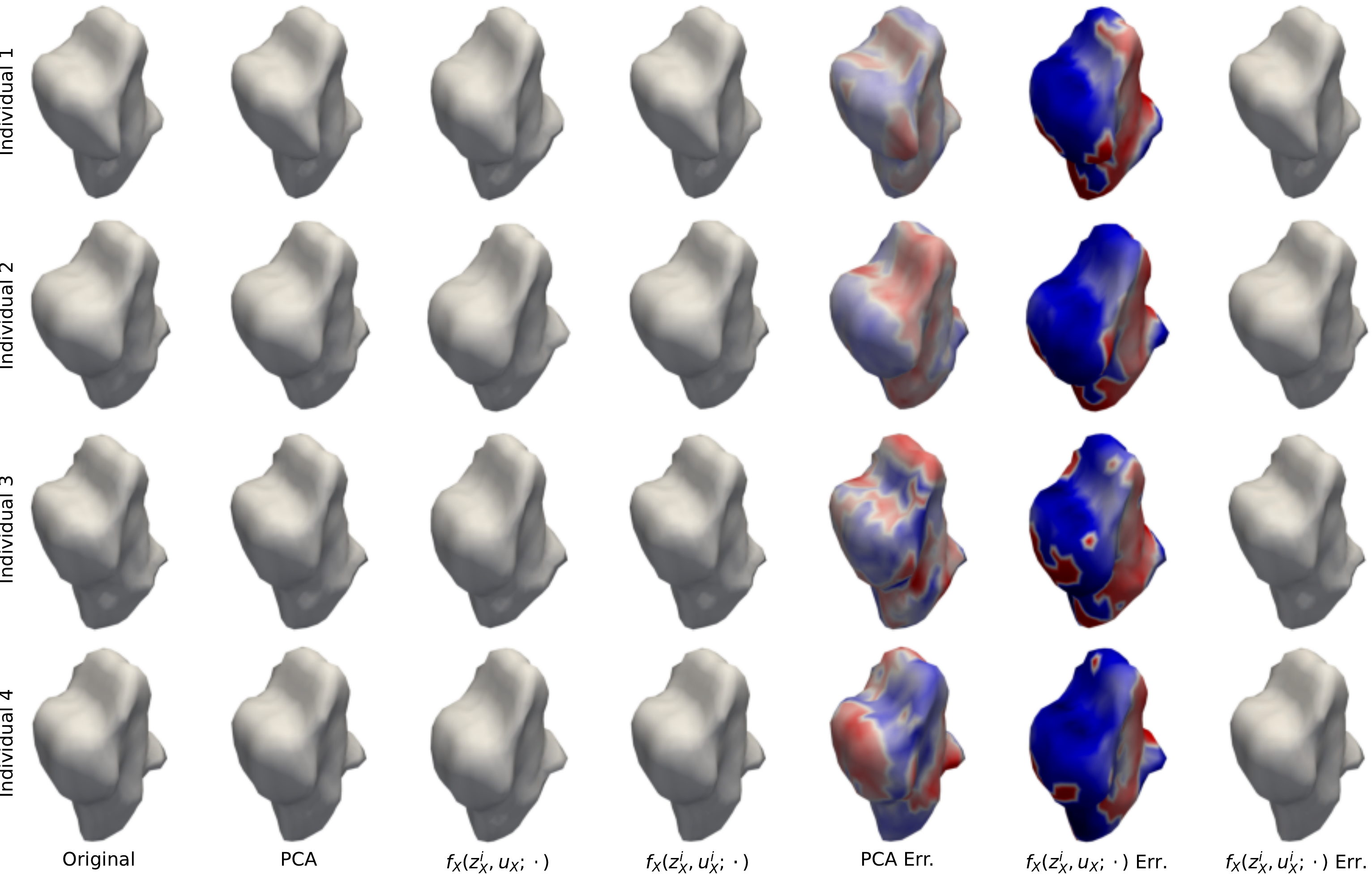}
        \caption{Reconstructions for 4 individuals produced by a PCA model and $f_X(
        \cdot)$. For PCA Err., \protect\markertwo\, = +3mm to \protect\markerone\, = -3mm. For $f_X(\cdot)$ Err., \protect\markertwo\, = +5mm to \protect\markerone\, = -5mm.}
        \label{fig:pca_vs_dscm_reconstruction}
    \end{center}
\end{figure}

\paragraph{\bf Generalisation.}
$f_X(\cdot)$'s ability to generalise unseen $\epsilon_X$ is quantified by the vertex Euclidean distance (VED) between an input mesh $x$ and its reconstruction $\tilde{x} \approx f_X(f_X\inv(x; v, b); v, b)$. Reconstruction VEDs are also an indication of the quality of the counterfactuals that can be generated by a CSM, since counterfactuals preserve vertex-level details from the initial observation \cite{pawlowski2020dscm}. To decouple the contributions of \noisehix\ and \noiselox, we compare reconstructions when $u_X \sim \mathcal{N}(0, I_{3|V|})$, $\tilde{x}^i = f_X(z_X^i, u_X; \cdot)$, against $u_X^i$ inferred for an individual $i$, $\tilde{x}^i = f_X(z_X^i, u_X^i; \cdot)$. PCA \cite{pca2006bishop} is used as a linear baseline model for comparisons.

\Cref{fig:pca_vs_dscm_reconstruction} visualises the reconstructions for four individuals with further results in \cref{app:experiments}. PCA with 32 modes (to match the dimensionality of the CSM) produces good quality reconstructions on average, albeit with spatially inconsistent errors across the surface meshes. Reconstructions from $f_X(z_X^i, u_X; \cdot)$ are spatially coherent, although lower quality (higher VED) on average; shapes are uniformly larger over the pons (anterior) and under the 4th ventricle (posterior), and uniformly smaller around the medulla (lateral). The exogenous variable $z_X$ therefore encodes the full range of shapes, which, when including a subject-specific $u_X^i$, $f_X(z_X^i, u_X^i; \cdot)$, captures scale with zero reconstruction error by construction. 

\begin{figure}[t]
    \centering
    \label{fig:image2}
    \begin{subfigure}{0.4\textwidth}
        \begin{center}
            \includegraphics[width=\linewidth]{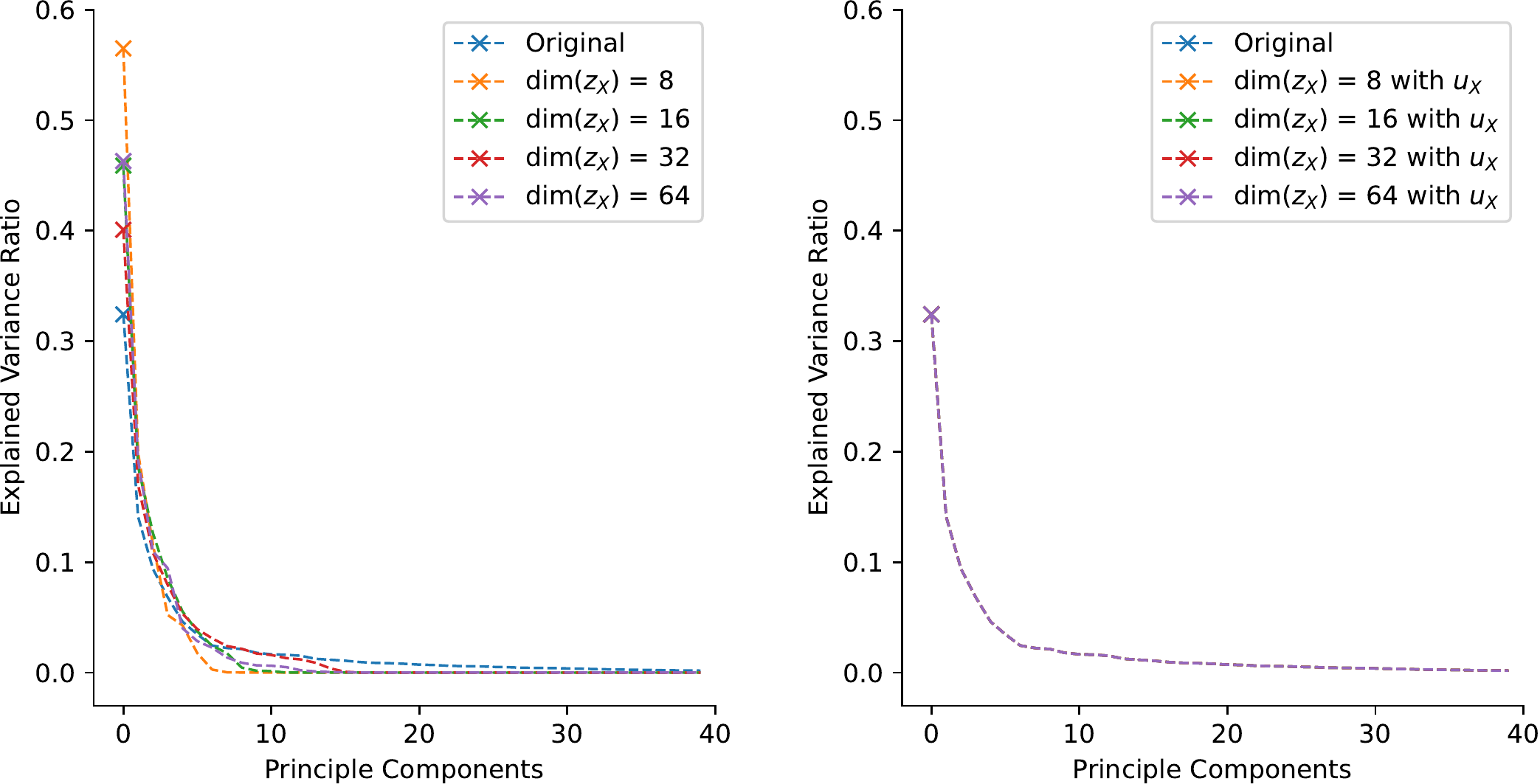}
            \caption{$u_X \sim \mathcal{N}(0, I_{3|V|})$ (\textbf{left}) and $u_X$ inferred for each individual $i$, $u_X^i$ (\textbf{right}).}
            \label{fig:reconstruction_compactness}
        \end{center}
    \end{subfigure}
    \begin{subfigure}{0.4\textwidth}
        \begin{center}
            \includegraphics[width=\linewidth]{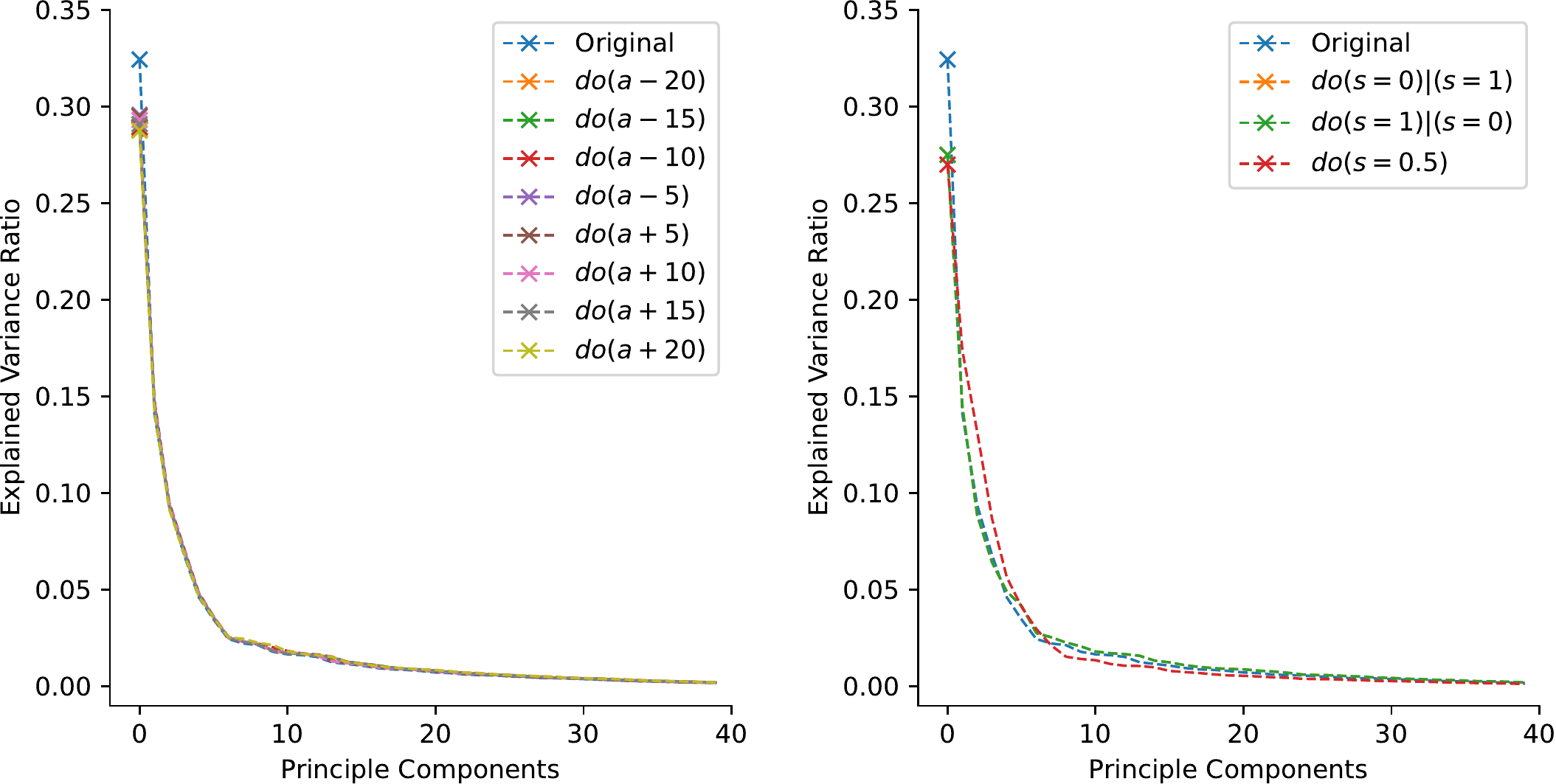}
            \caption{Interventions on age $do(a)$ (\textbf{left}) and sex $do(s)$ (\textbf{right}).}
            \label{fig:counterfactual_compactness}
        \end{center}
    \end{subfigure}
    \caption{Explained variance ratio of PCA models fitted to mesh reconstruction (\cref{fig:reconstruction_compactness}) and counterfactuals (\cref{fig:counterfactual_compactness}) compared to PCA fitted to the original shapes.}
\end{figure}

\paragraph{\bf Reconstructed Shape Compactness.}
Here, PCA is used on reconstructions from the test set to measure compactness compared to PCA on the original shapes. The explained variance ratio, $\lambda_k / \sum_{j} \lambda_j$, for each eigenmode $\lambda_k$ across all PCA models is visualised in \cref{fig:reconstruction_compactness}. By overfitting the principal mode in the true distribution, corresponding to the axis along which the majority of the mesh's volume is concentrated, the CSM with $D = 8$ overfits the objective. There is a large spread in the per-vertex error, although the average mesh VED is low. On the other hand, the CSM with $D = 32$ has a primary explained variance ratio closer to the original and its curve plateaus more gradually, suggesting that reconstructions preserve a wider range of shapes. The results in \cref{sec:individual_specific_shapes}, provide further explanations for how compactness affects the quality of counterfactuals.

\begin{figure}[b]
    \begin{center}
        \includegraphics[width=.6\textwidth]{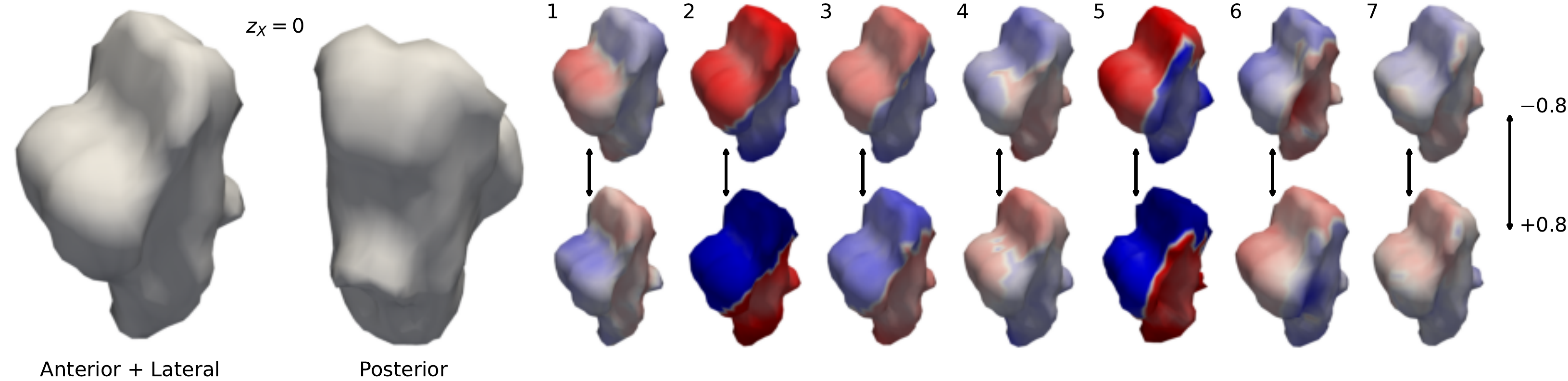}
        \caption{Interpolating the latent space $z_X$. The mean shape is on the left, followed by $(\bar{z_X})_k \pm 0.8$ for $k \in [1, 7]$, the first 7 components, to the right.}
        \label{fig:shape_interpolation_population_level_simplified}
    \end{center}
\end{figure}

\subsection{Interventional Level: Population-Level Shape Generation}
\label{sec:interventional_results}

\paragraph{\bf Interpolating the Exogenous Shape Space $z_X$.}
The CSM's mean brain stem shape $\bar{x}$ is generated by setting $\text{\noisehix} = \bar{\text{\noisehix}} = 0$ and sampling the mesh generating mechanism as $\bar{x} = f_X(\bar{\text{\noisehix}}, u_X; \bar{b}, \bar{v})$, where $u_X \sim \mathcal{N}(0, I_{3|V|})$, and $\bar{b}$ and $\bar{v}$ are the $b$ and $v$ means, respectively. We vary \noisehi\, by interpolating the $k^\text{th}$ dimension, $(\bar{z_{X}})_k$, as $(\bar{\text{\noisehix}})_k + j c$, whilst all other dimensions are fixed at 0. In \cref{fig:shape_interpolation_population_level_simplified}, $j \in \{-4, 4\}$, $c = 0.2$ and the mean face is on the left. Each latent dimensions is responsible for precise shapes, since the deformed regions are separated by a fixed boundary where VED = 0mm. In order to preserve $\bar{b}$ and $\bar{v}$, a contraction on one side of the boundary corresponds to an expansion on the other side.

\paragraph{\bf t-SNE Shape Projection.}
We can understand how parent, and \textit{grandparent}, causes affect brain stem shapes across the population by projecting (near-)mean meshes, under interventions, onto a low-dimensional manifold. We choose a shape representative of the population, $z_X^i \approx 0$, sample $a^k$ and $s^k$ from the learned distributions, and then perform the intervention $do(a^k, s^k)$. Post-interventional meshes are sampled as: (1)~$b^k = f_B(\epsilon_B^k; a^k, s^k)$; (2)~$v^k = f_V(\epsilon_V^k; a^k, b^k)$; (3)~$x^{ik} = f_X(z_X^i, u_X^k; v^k, b^k)$. By projecting $\{x^{ik}\}_{k=1}^{5000}$ using t-SNE \cite{van2008visualizing}, we can visualise learned correlations between meshes and their causes.

\begin{figure}[tb]
\centering
\begin{minipage}{.5\textwidth}
    \centering
    \includegraphics[width=\textwidth]{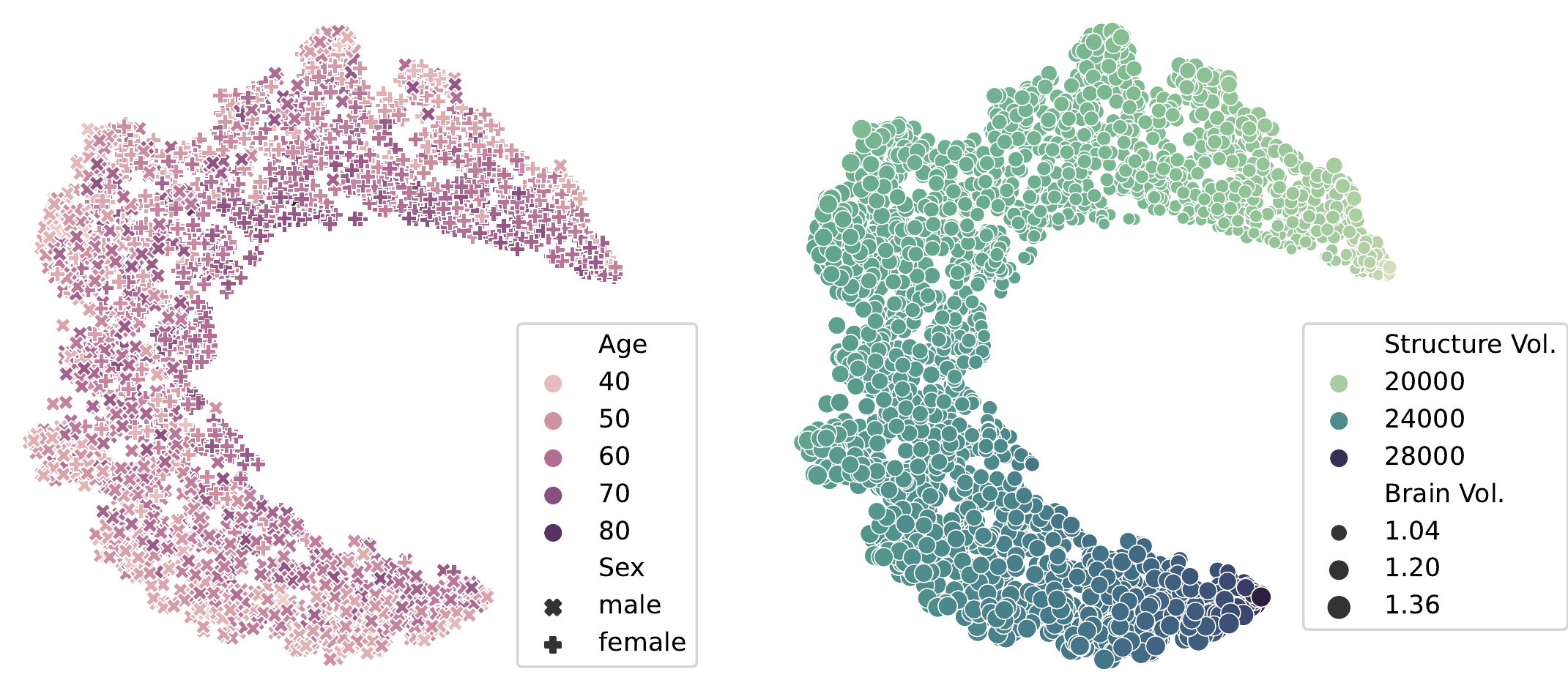}
    \captionof{figure}{t-SNE \cite{van2008visualizing} projection of the meshes in $\{x^{ik}\}$.}
    \label{fig:shape_projection_population_level}
\end{minipage}
\begin{minipage}{.41\textwidth}
    \centering
    \captionof{table}{Average specificity errors of the set $\{e^k\}$ in millimetres (mm).}
    \label{tab:specificity}
    \begin{tabular}{lcc}
        \toprule
        $do(\cdot)$ & Mean $\pm$ Error & Median \\
        \midrule
        $do(a, s)$    & 1.246 $\pm$ 0.128 & 1.198 \\
        $do(b, v)$    & 2.034 $\pm$ 0.680 & 1.832 \\
        \bottomrule
    \end{tabular}
\end{minipage}
\end{figure}

In \cref{fig:shape_projection_population_level} (right), we see that brain stems have ($b$, $v$)-related structures, which are correlated to $a$ and $s$ (left). These correlations are causal, since $a$ and $s$ are parents of $b$ and $v$ in the causal graph. Meshes also present trends from the true distribution; females and older people are associated with smaller $b$ and $v$ values, the male and female clusters are similar in size, males span a larger range of volumes. Notice also that a setting of $(a^k, s^k, b^k, v^k)$ is associated with a variety of mesh shapes, due to random sampling of $u_X$. We can conclude that $(a, s)$-related trends can manifest themselves in volumes, to discriminate sex, and shapes, to regress over age. 

\paragraph{\bf Specificity of Population-Level Shapes.}
The specificity of the population-level model is used to validate brain stem mesh shapes produced under a range of interventions. We generate 100 meshes for each setting of the interventions $do(a, s)$ and $do(v, b)$ by repeatedly sampling $u_X$. $z_X$ is the same as that used to produce \cref{fig:shape_projection_population_level}. For each generated mesh $x^k$, we calculate \\ $e^k =\frac{1}{|X_{test}|}$ $\sum_{x^t \in X_{test}}\mathrm{VED}(x^k, x^t)$, where $X_{test}$ is the set of test meshes. The mean and median of the set $\{e^k\}$ define the population-level \textit{specificity errors}, and are presented in \cref{tab:specificity}. These results are on par with the generalisation errors in \cref{app:experiments} and \cref{fig:individual_traits_bars}, demonstrating quantitatively that realistic meshes are produced under the full range of interventions.

\subsection{Counterfactual Level: Subject-Specific Shape Generation}
\label{sec:individual_specific_shapes}

\begin{figure}[t]
    \begin{center}
        \includegraphics[width=.7\textwidth]{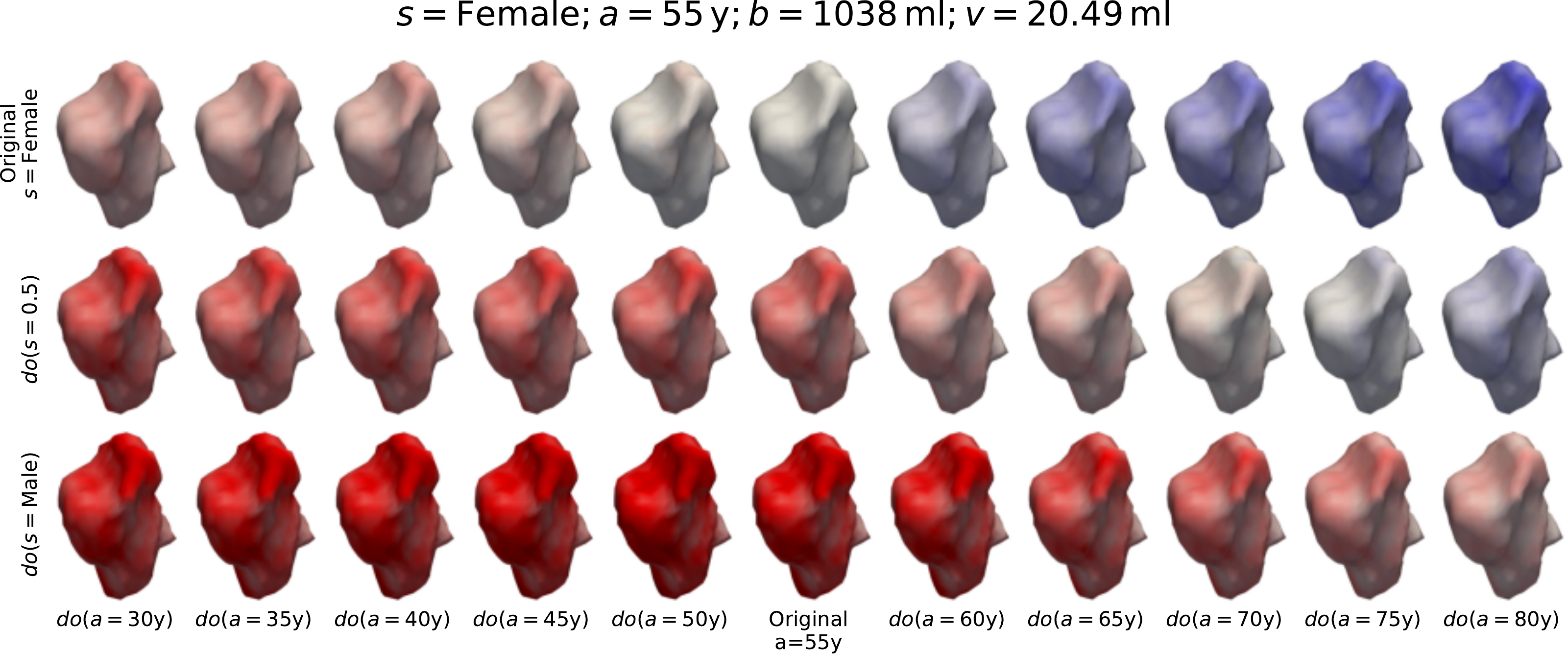}
    \end{center}
    \caption{Counterfactual meshes for an individual under $do(a)$ and $do(s)$ -- \textit{``What would this person's brain stem look like if they were older/younger or male?"}. Colours show the VED between observed and counterfactual meshes -- \protect\markertwo\, = +5mm to \protect\markerone\, = -5mm.}
    \label{fig:counterfactual_interpolation_main}
\end{figure}

\paragraph{\bf Interpolating and Extrapolating Causes.}
In \cref{fig:counterfactual_interpolation_main}, we visualise counterfactual meshes under interventions $do(a)$ (rows), $do(s)$ (columns) and $do(a, s)$. Our results qualitatively demonstrate that counterfactuals generated by the CSM generalise to interventions on unseen values, e.g. $do(s = 0.5)$ or $do(a = 80\text{y})$, since subject-specific features are preserved whilst population-level trends are present. Namely, volume decreases as age increases, a male brain stem is larger than its female counterpart for the same age by a constant volume factor, and counterfactual meshes for $do(s = 0.5)$ are \textit{half way} between the male and female for the same age. Notice also that mesh deformations are smooth as we interpolate age and sex, without straying towards the population-level mean shape ($z_X = 0$ in \cref{fig:shape_interpolation_population_level_simplified}).

\begin{figure}[t]
    \begin{center}
        \includegraphics[width=.6\textwidth]{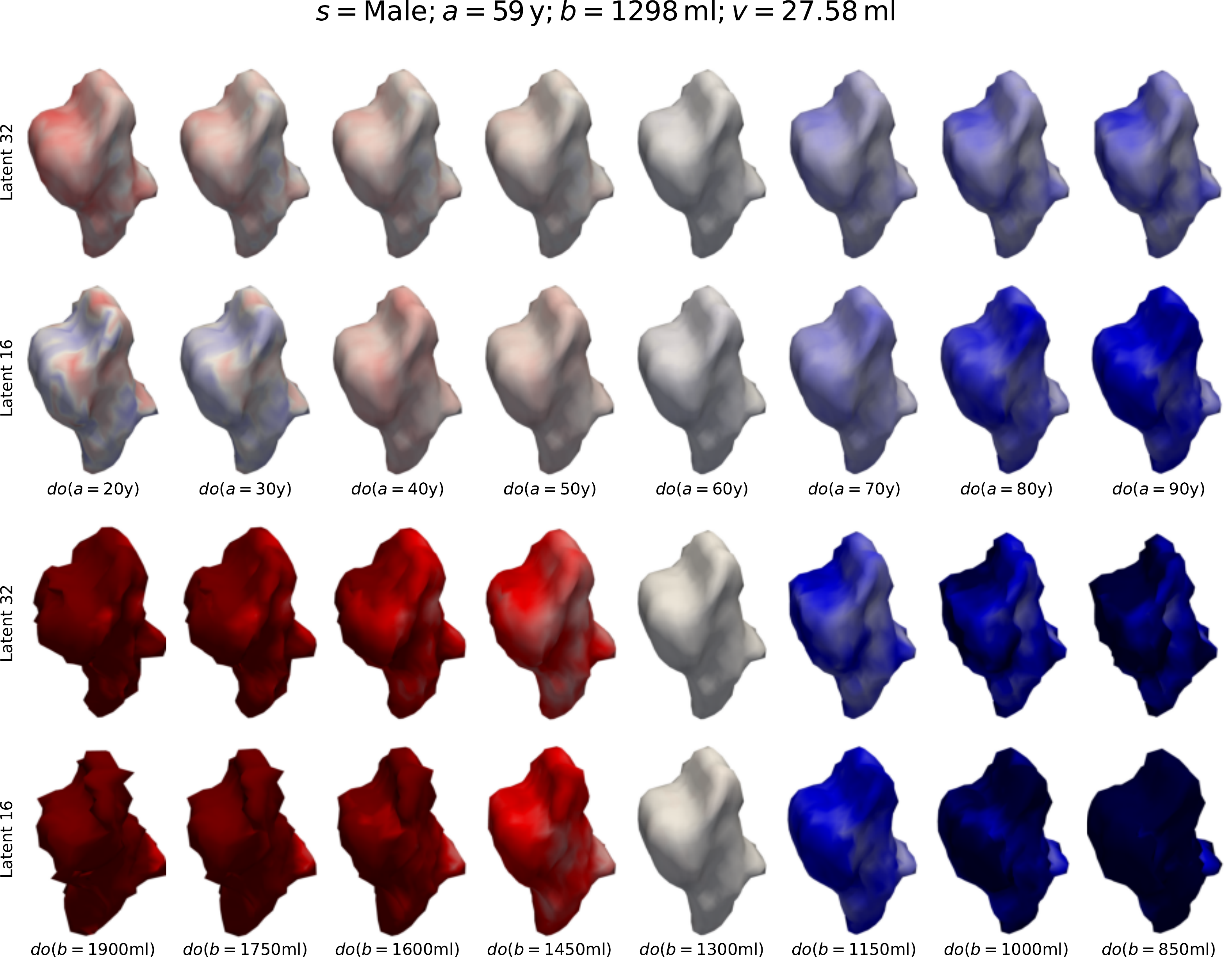}
        \caption{Counterfactual brain stem meshes for an individual, under $do(a)$ and $do(b)$ interventions, produced by DSCMs with different values for $D$. Colours show the VED between the observed and counterfactual meshes -- \protect\markertwo\, = +5mm to \protect\markerone\, = -5mm.}
        \label{fig:counterfactual_interpolation_latent_comparison}
    \end{center}
\end{figure}

In \cref{fig:counterfactual_interpolation_latent_comparison}, we compare counterfactual meshes produced by CSMs with $D \in \{16, 32\}$ under $do(a)$ and $do(b)$. Both CSMs generate meshes that adhere to the population-level trends but differ significantly under interventions at the fringes of the $a$ and $b$ distributions, e.g. $do(b = 1900 \text{ml})$, $do(a = 20 \text{y})$ or $do(a = 90 \text{y})$. When $D = 16$, each dimension in $z_X$ corresponds to a combination of shape features, resulting in uniform or unsmooth deformations. This can be explained by plots in \cref{fig:reconstruction_compactness}, where variance for $D = 16$ is concentrated in the principle component, suggesting that many large shape deformations are correlated. On the other hand, reconstructions by $D = 32$ are less compact implying that deformations occur \textit{more} independently, resulting in higher quality mesh counterfactuals. 

\begin{figure}[t]
    \begin{center}
    \begin{subfigure}{.7\textwidth}
        \includegraphics[width=\textwidth]{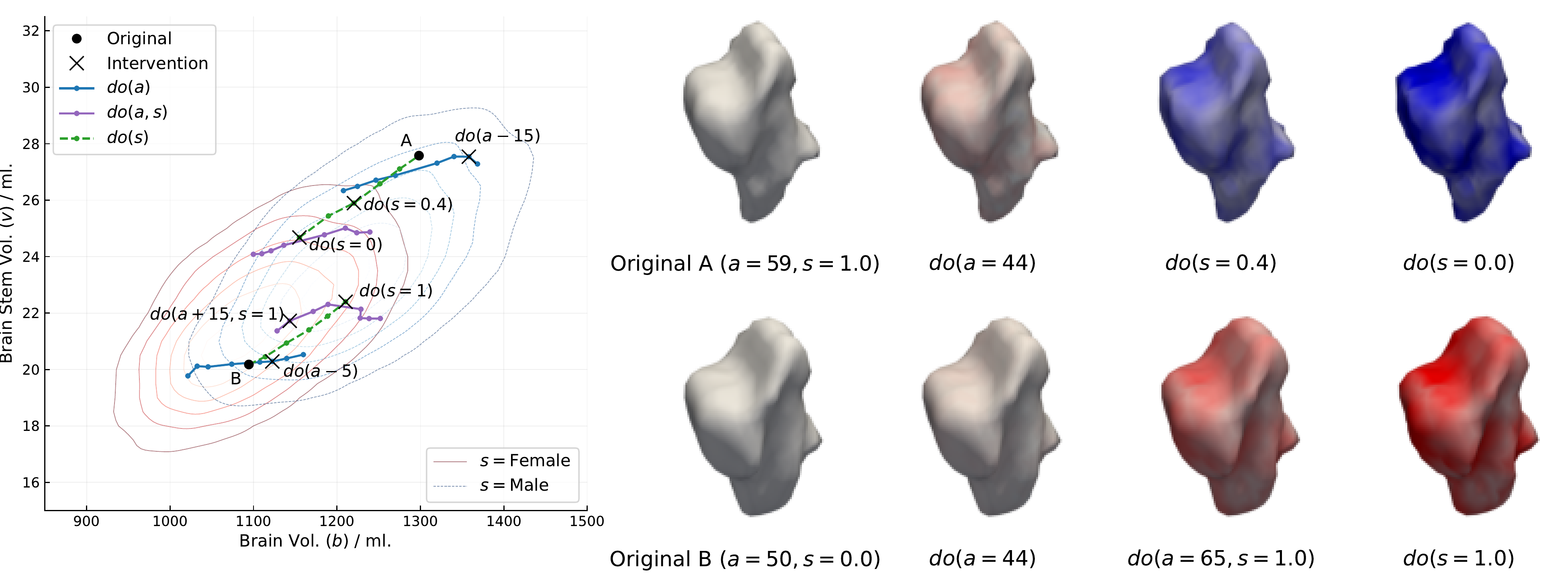}
    \end{subfigure}
    \begin{subfigure}{.7\textwidth}
        \includegraphics[width=\textwidth]{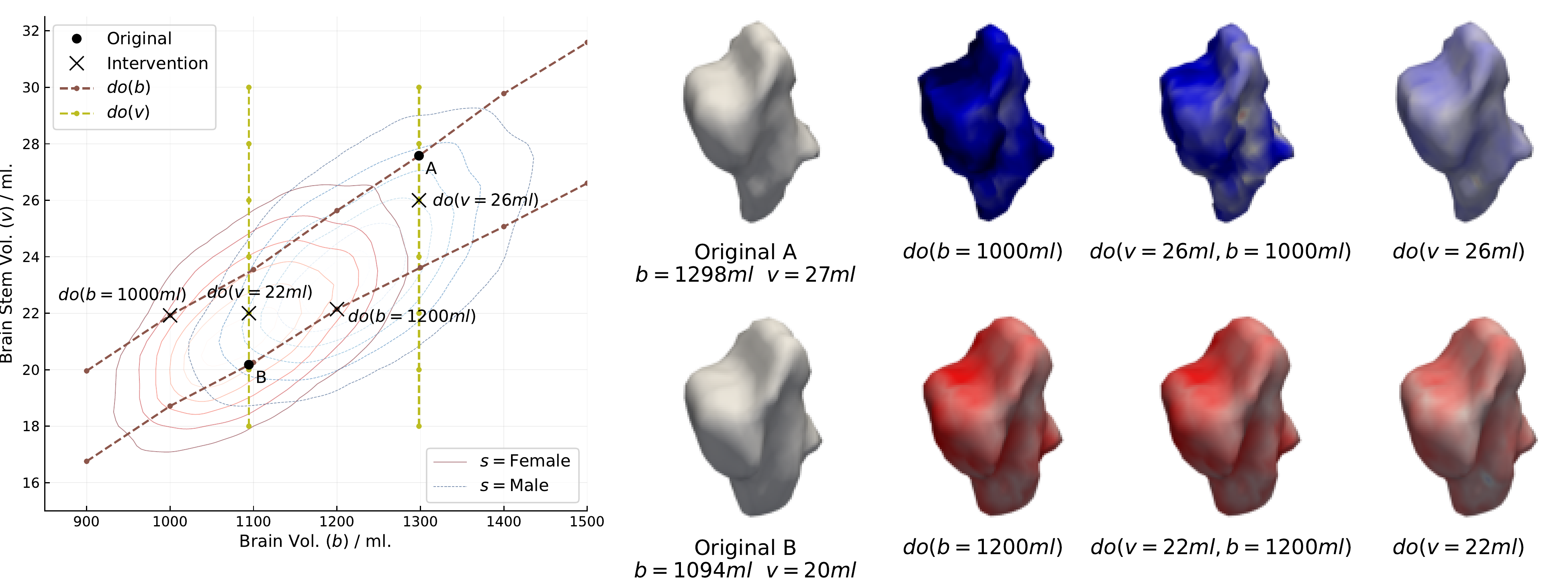}
    \end{subfigure}
    \end{center}
    \caption{Counterfactual trajectories of $b_\cf$ and $v_\cf$ under interventions $do(a)$, $do(s)$ and $do(a, s)$ (\textbf{top}), and $do(b)$ and $do(v)$ (\textbf{bottom}). Trajectories are plotted on top of the contour plot for the density $p(v,b|s)$. Colours show the VED between the observed and counterfactual meshes -- \protect\markertwo\, = +5mm to \protect\markerone\, = -5mm.}
    \label{fig:trajectories}
\end{figure}

\paragraph{\bf Counterfactual Trajectories.}
In \cref{fig:trajectories} (top), we plot the (b, v)-trajectories under interventions $do(a \pm T)$, $do(s = S)$, and $do(a \pm T, s = S')$, where $T \in \{5, 10, 15, 20\}$, $S \in \{0, 0.2, 0.4, 0.6, 1\}$ and $S'$ is the opposite of the observed sex. \cref{fig:trajectories} (bottom) shows $do(b)$ and $do(v)$. Counterfactuals, corresponding to marked ($\times$) points on the trajectories, are presented on the right in both plots.

The trends in the true distribution are present in the counterfactual trajectories for persons A and B, with some subject-specific variations. For example, $do(s=0)$ shifts $v$ and $b$ to the female region of the distribution for person A. This causes an overall shrinkage of A's brain stem, from a complex, non-uniform transformation of the mesh surface. The post-interventional distribution of $v$ under $do(b)$ collapses to its conditional distribution, $p(v|do(b)) = p(v|b)$, since we assume $b \rightarrow v$ in the causal graph. On the other hand, intervening on $v$ fixes the output of $f_V(\cdot)$ and does not affect $b$, resulting in vertical trajectories. Notice also that for person B, the following counterfactual meshes have similar shapes, as per the trajectories: $f_X(\epsilon_X; f_V(\epsilon_V; a, 1200\text{ml}), 1200\text{ml})$ = $f_X(\epsilon_X; 22\text{ml}, 1200\text{ml})$ = $f_X(\epsilon_X; f_V(\epsilon_V; a, b_\cf), b_\cf)$, where $b_\cf = f_B(\epsilon_B; a, \text{Male})$. In \cref{fig:counterfactual_interpolation_latent_comparison} and \cref{fig:trajectories}, counterfactuals for person A become increasingly unrealistic under large changes to $b$. This is not the case, however, when age interventions are the cause of these changes. This is due to the usage of the constrained, latent variables $\hat{b}$ and $\hat{v}$ (\cref{sec:dscm_architecture}), ensuring that $f_X(\cdot)$ can generate high-quality meshes under out-of-distribution interventions. Furthermore, the inferred $\epsilon_B$ and $\epsilon_V$ during abduction prevent $do(a)$ from causing unrealistic $b_\cf$ and $v_\cf$ values for an individual.

\paragraph{\bf Counterfactual Compactness.}
In \cref{fig:counterfactual_compactness}, the explained variance ratio is lower for counterfactuals than observations across all age and sex interventions, suggesting that counterfactuals have a smaller range of volumes. All modes beyond the principle component explain the same ratio of variance for both, demonstrating that $f_X(\cdot)$ is able to preserve the full range of subject-specific shapes during counterfactual inference, regardless of differences in scale. 

\paragraph{\bf Preservation of Subject-Specific Traits.}
Here, we demonstrate quantitatively that our CSM can decouple $\epsilon_X$ and $pa_X$ in $f_X(\cdot)$ and $f_X\inv(\cdot)$. Consider the observations ($x^i, a^i, s^i, b^i, v^i$) for an individual $i$ and generate a counterfactual mesh $x_\cf^i$ under an intervention on age, $do(a=A)$ as:  (1)~$b_\cf^i = f_B(\epsilon_B^i; A, s^i)$; (2)~$v_\cf^i = f_V(\epsilon_V^i; A, b_\cf^i)$; (3)~$x_\cf^i = f_X(z_X^i, u_X^i; v_\cf^i, b_\cf^i)$, where ($z_X^i$, $u_X^i$, $\epsilon_V^i$, $\epsilon_B^i$, $\epsilon_A^i$) are inferred during the abduction step. We then reconstruct the observed mesh from the counterfactual by counterfactual inference. To do this, we start with ($x_\cf^i, A, s^i, b_\cf^i, v_\cf^i$) and generate counterfactuals on $do(b=b^i)$ and $do(v=v^i)$ as ${x_\cf^i}' = f_X({z_X^i}', {u_X^i}'; v^i, b^i)$, where ${\epsilon_X^i}' = ({z_X^i}', {u_X^i}')$ are inferred during an abduction step. Provided that $z_X^i = {z_X^i}'$ and $u_X^i = {u_X^i}'$, we expect
\begin{align}
    {x_\cf^i}' = f_X({z_X^i}', {u_X^i}'; v^i, b^i) = f_X(z_X^i, u_X^i; v^i, b^i) = x^i.
    \label{eq:counterfactual_reconstructions_equivalence}
\end{align}
These steps are also performed for interventions on sex, $do(s)$. Since $f_X(\cdot)$ is implemented as an amortised, explicit mechanism, we expect that ${x_\cf^i}' \approx x^i$ and ${\epsilon_X^i}' \approx \epsilon_X^i$ in practice.

\begin{figure}[t]
    \begin{center}
        \includegraphics[width=.6\textwidth]{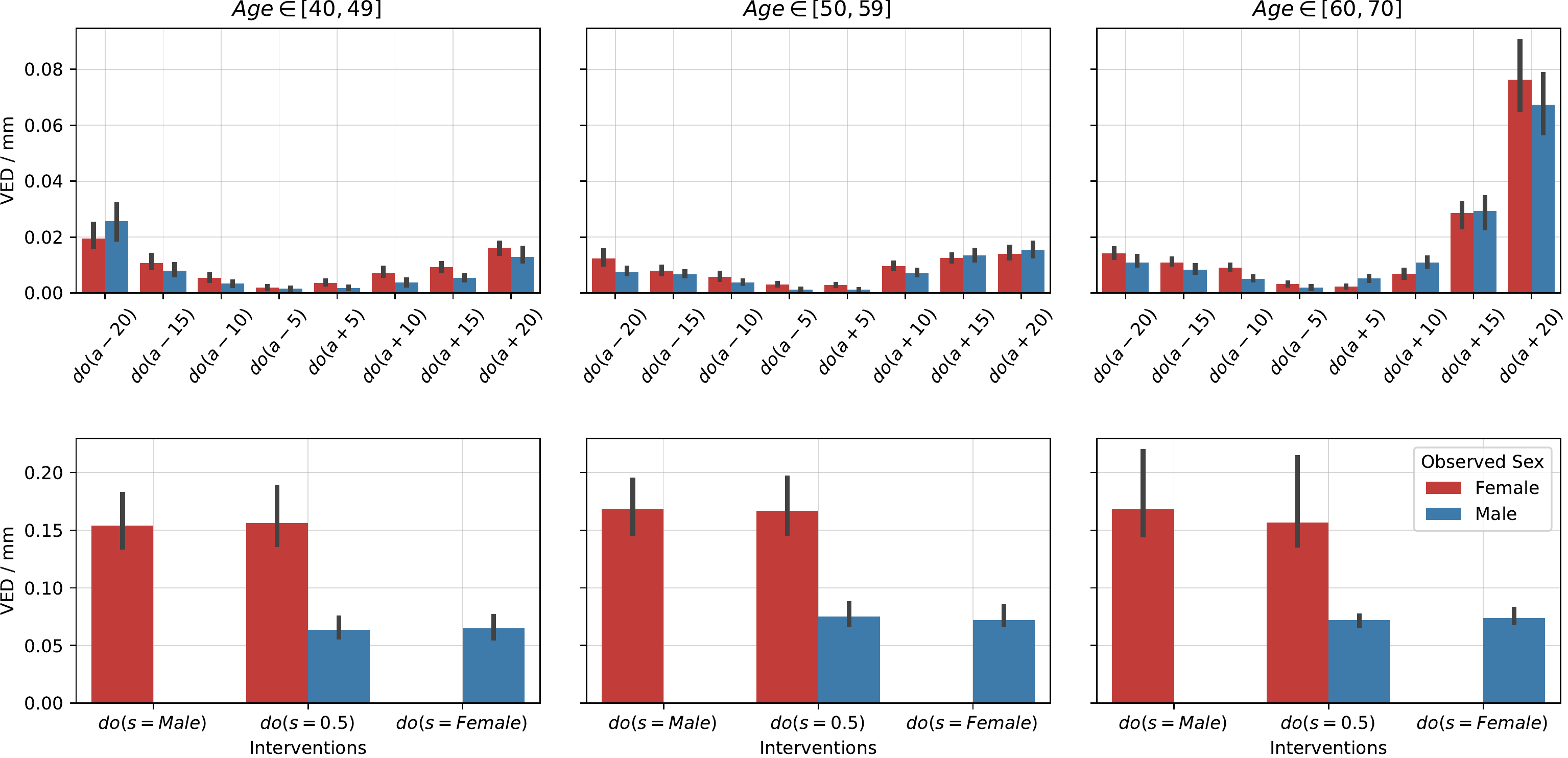}
        \caption{Preservation of subject-specific traits under counterfactual inference. Bars indicate the median of the VEDs between ${x_\cf^i}'$ and $x^i$ (\cref{eq:counterfactual_reconstructions_equivalence}).}
        \label{fig:individual_traits_bars}
    \end{center}
\end{figure}

\begin{figure}[b]
    \begin{center}
        \includegraphics[width=0.5\textwidth]{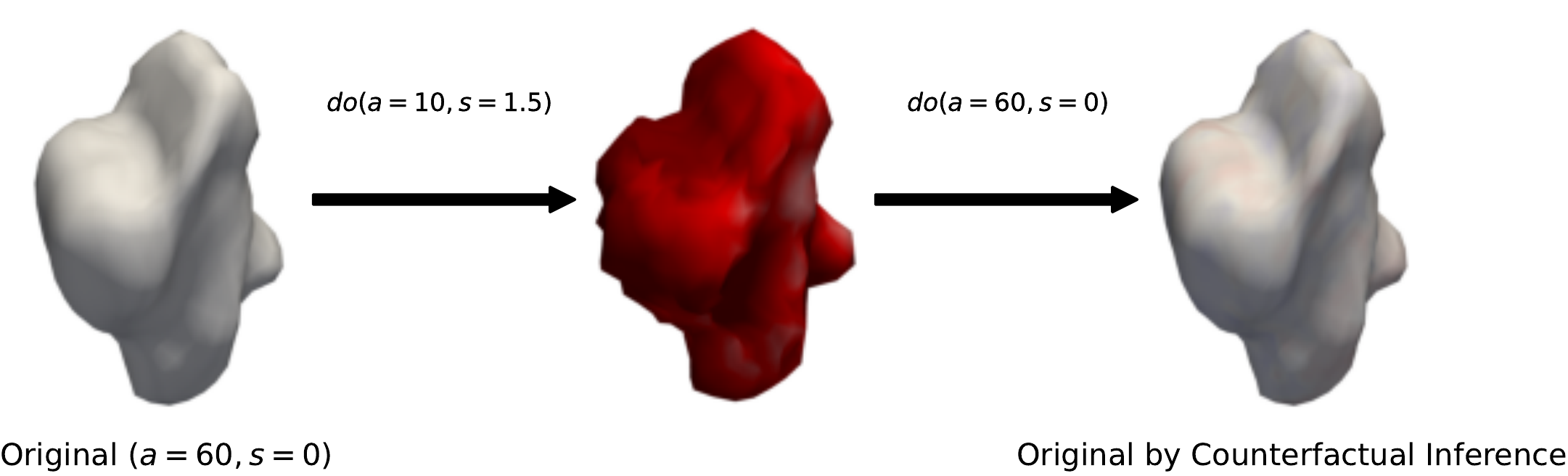}
        \caption{Reconstruction of the initial mesh under counterfactual inference.}
        \label{fig:individual_traits_recovery}
    \end{center}
\end{figure}

The median of the VEDs between ${x_\cf^i}'$ and $x^i$ is presented in \cref{fig:individual_traits_bars}. The CSM has learned high-quality mappings which can disentangle $\epsilon_X$ from its $x$'s causes, since we can recover meshes $x^i$ to within 0.2mm over the full range of interventions. This is seen in \cref{fig:individual_traits_recovery}, where $x_\cf$ is generated under an \textit{extreme} intervention and recovered successfully. The accuracy of the recovery depends on how far away from a variable's mode an intervention shifts the observed value. For example, $do(a^i + A')$ where $A' \in \{15, 20\}$ and $a^i \in [60, 70]$ results in $x_\cf'$ with the highest median VED, since $a^i + A' \in [65, 90]$ is generally outside the learned range of ages. Notice also that VEDs are lower for males than females when intervening on sex. In \cref{fig:shape_projection_population_level}, male brain stems are associated with a greater range of volumes and shapes. Therefore, going \textit{back to} the original sex, which causes a large change in volume, may generalise better for males than females.

\section{Conclusion}
In this paper, we presented Deep Structural Causal Shape Models capable of subject-specific and population-level mesh inference. Our model was able to generate novel counterfactual meshes which were robust to out-of-distribution interventions, preserved subject-specific traits, and captured the full range of shapes in the true data distribution. We demonstrated these capabilities by performing subject-specific shape regression in a biomedical scenario, without loss of generality, where longitudinal data is notoriously difficult to collect. Consequently, we developed a mesh CVAE, with geometric deep learning components, capable of high-quality brain structure mesh generation. In doing so, we considered the challenges involved with conditional shape modelling using spectral convolutions and incorporating a mesh CVAE into a DSCM. Due to the modularity of our proposed CSMs, latest techniques in generative 3D shape modelling can be easily integrated to further improve the quality of our results. In future work, we intend to build a library for mesh generation, akin to \cite{castro2019morpho}, to model the assumed data generation process and generate reference counterfactuals.

\paragraph{\textbf{Acknowledgements.}}
This project has received funding from the European Research Council (ERC under the European Union’s Horizon 2020 research and innovation programme (Grant Agreement No. 757173, Project MIRA). The UK Biobank data is accessed under Application Number 12579.

\clearpage

\bibliographystyle{splncs04}
\bibliography{bibliography}

\clearpage

\input{appendix}
\clearpage

\end{document}

%% file: tikz/sem.tex
\newsavebox\submedical
\sbox{\submedical}{%
\begin{tikzpicture}[x=8mm, y=6mm]
  \node[cvar, label={left:$a$}] at (0,2) (va) {};
  \node[cvar, label={right:$s$}]at (1,2) (vs) {};
  \node[cvar, label={left:$v$}] at (0,1) (vv) {};
  \node[cvar, label={right:$b$}]at (1,1) (vb) {};
  \node[cvar, label={right:$x$}]at (.5,0) (vx) {};
  
  \edge{va,vs}{vb};
  \edge{va,vb}{vv};
  \edge{vv,vb}{vx};
\end{tikzpicture}
}
\begin{tikzpicture}
  \node[latent]                            (ea) {$\epsilon_A$};
  \node[obs, right=of ea]         (a) {$a$};
  
  \node[latent, right=2 of a]               (es) {$\epsilon_S$};
  \node[obs, left=of es]                  (s) {$s$};
  
  \node[obs, below=of a]                  (v) {$v$};
  \node[latent, left=of v]               (ev) {$\epsilon_V$};
  
  \node[obs, below=of s]                  (b) {$b$};
  \node[latent, right=of b]               (eb) {$\epsilon_B$};

  \node[aux, below=0.5 of a] (hv) {};
  \node[aux, below=0.5 of s] (hb) {};
  
  \node[obs] at ($(v)+(.5,-1.25)$)   (x) {$x$};
  \node[latent, right=of x]               (exlo) {$u_X$};
  \node[latent, above=.75 of exlo]          (exhi) {$z_X$};
  \node[aux, below=.5 of b] (hx) {};
  \node[aux, above=.5 of x] (kx) {};
  
  \edge[invertible]{ea}{a}
  \edge{es}{s}
  \draw[invertible] (ev) -- (v) coordinate[midway] (fv);
  \draw[invertible] (eb) -- (b) coordinate[midway] (fb);
  \edge[-]{a, b}{hv}
  \edge[-]{a, s}{hb}
  \draw[-Circle] (hv) edge[out=202.5, in=90] (fv);
  \draw[-Circle] (hb) edge[out=-22.5, in=90] (fb);
  \edge[-]{v, b, exhi}{hx}
  \draw[invertible] (exlo) -- (x) coordinate[midway] (fx);
  \edge[-Circle]{hx}{fx}

  \edge[amort, -]{v, b, x}{kx};
  \draw[amort] (kx) edge[->, bend right=30] (exhi);
  
  \node[fill=gray!10, inner sep=2pt, scale=.8] at ($(x)+(-1.2,.15)$) {\usebox{\submedical}};
\end{tikzpicture}

%% file: appendix.tex
\appendix

\section{Additional Background Material}
\label{sec:additional_background}

\subsection{DSCM Mechanisms}
We briefly outline mechanism abstractions in the DSCM framework \cite{pawlowski2020dscm} relevant to this work.

\paragraph{\bf Invertible, explicit}
This mechanism is used to generate low-dimensional variables. For assignments of the form $x_m := f_m(\epsilon_m)$, $f_m(\cdot)$ is implemented using a normalising flow \cite{papamakarios2021normalizing} and $\epsilon_m$ is sampled from a standard normal. The flow's parameters are optimised by maximising the exact, explicit likelihood of $p(x_m)$,
\begin{gather}
    p(x_m) = p(\epsilon_m) \cdot |\text{det} \: \nabla_{\epsilon_m} f_m(\epsilon_m)|\inv,
\end{gather}
where $\epsilon_m = f_m\inv(x_m)$ is computed by inverting the flow. When modelling mechanisms of the form $x_m := f_m(\epsilon_m; pa_m)$, $f_m(\: \cdot \:; \: pa_m)$ is implemented using conditional normalising flows \cite{winkler2019learning}. The explicit likelihood for $x_m$ given its causes $pa_m$ is now
\begin{gather}
    p(x_m|\; pa_m) = p(\epsilon_m) \cdot |\text{det} \: \nabla_{e_m} f_m(\epsilon_m; pa_m)|\inv,
\end{gather}
where $\epsilon_m = f_m\inv(x_m; pa_m)$. Note that for complex distributions, the flow can include chains of simple, diffeomorphic functions, such as sigmoids and exponentials. \\

\paragraph{\bf Amortised, explicit}
This mechanism is used to generate high-dimensional data $x_m$, and efficiently approximate $\epsilon_m$'s during abduction. An assignment $x_m := f_m(\epsilon_m; pa_m)$ is implemented as 
\begin{gather}
    x_m := f_m(\epsilon_m; pa_m) = l_m(u_m; h_m(z_m; pa_m), pa_m),
    \label{eq:amortized_explicit_assignment}
\end{gather}
where $\epsilon_m = (u_m, z_m)$, $p(u_m)$ is an isotropic Gaussian, $p(z_m|x_m, pa_m)$ is approximated by the variational distribution $q(z_m|x_m, pa_m)$, $l_m(u_m; w_m)$ implements a simple location-scale transformation, $u_m \odot \sigma(w_m) + \mu(w_m)$, and $h_m(z_m; pa_m)$ parametrises $p(x_m|\, z_m, pa_m)$ using a conditional decoder, $\text{CondDec}_{m}(z_m; pa_m)$. The decoder's parameters are jointly optimised with a conditional encoder, $\text{CondEnc}_{m}(x_m; pa_m)$, which parametrises $q(z_m|x_m, pa_m)$, by maximising the evidence lower bound (ELBO) on $\log p(x_m|pa_m)$,
\begin{align}
    \log p(x_m |\; pa_m) &\geq E_{q(z_m|x_m, pa_m)}[\log p(x_m, z_m, pa_m) - \log q(z_m | x_m, pa_m)], \label{eq:elbo_example} \\
    &= E_{q(z_m|x_m, pa_m)}[\log p(x_m | z_m, pa_m)] - \text{KL}[q(z_m | x_m, pa_m)\|p(z_m)], \notag 
\end{align}
where the density $p(x_m|z_m,pa_m)$ is calculated as
\begin{align}
    p(x_m| z_m, pa_m) = p(u_m) \cdot |\text{det} \: \nabla_{u_m} l_m(u_m; h_m(z_m; pa_m), pa_m)|\inv,
\end{align}
with $u_m = l_m\inv(x_m; \: h_m(z_m; pa_m), pa_m)$. This forms a CVAE \cite{sohn2015learning} in the DSCM.

\subsection{Pearl's Causal Hierarchy}
\label{sec:pch}
Pearl's causal hierarchy \cite{pearl2019seven} is a framework for classifying statistical models by the questions that they can answer. At the first level, associational models can learn statistical relationships between observed variables i.e. $p(y|x)$. At the next level, interventions can be performed on the data generating process by fixing the output of a mechanism $f_m(\cdot)$ to $b$, denoted as $do(x_m := b)$ \cite{richardson2013single}. Finally, structural causal models (SCM) can be used to simulate subject-specific, retrospective, hypothetical scenarios, at the counterfactual level \cite{glymour2016causal}, as follows:
\begin{enumerate}
    \item Abduction - Compute the exogenous posterior $p(\mathcal{E} | X)$.
    \item Action - Perform any interventions, e.g. $do(x_j := a)$, thereby replacing the mechanism $f_j(\cdot)$ with $a$. The SCM becomes $\tilde{G} = (\mathcal{E}, p(\mathcal{E} | X), X, F_{do(x_j := a)})$.
    \item Prediction - Use $\tilde{G}$ to sample the counterfactual distribution $p_{\tilde{G}}(x_k|pa_k)$ using its associated mechanism $f_k(\cdot)$.
\end{enumerate}

\subsection{Counterfactual Inference using DSCM Framework}
In \cref{sec:pch}, counterfactual inference is presented in the general case of partial observability and non-invertible mechanisms. The DSCM framework instead assumed unconfoundedness and provides a formulations where $\epsilon_m$ can be approximately inferred. The previous process can therefore be modified to use the mechanisms directly, as outlined in \cref{tab:deep_counterfactual_inf}.
 
\begin{table}[b]
    \centering
    \caption{Steps for counterfactuals inference for different types of mechanisms $f_m(\cdot)$ in the DSCM framework given an observation $x_m$. Counterfactuals are denoted with the subscript `$\cf$'.}
    \label{tab:deep_counterfactual_inf}
    \setlength{\tabcolsep}{.5em}
    \scalebox{0.90}{
    \begin{tabular}{lcc}
        \toprule
         Step & Invertible-explicit & Amortised-explicit \\
         \midrule
         Abduction & 
         $\epsilon_m = f\inv_m(x_m; pa_m)$
         &
         \makecell{
            \begin{tabular}{c@{ }l}
            &$\epsilon_m = (z_m, u_m)$: \\
            (1) &$z_m \sim q(z_m|x_m, pa_m)$ \\
            (2) &$u_m = l_m\inv(x_m; h_m(z_m, pa_m), pa_m)$
            \end{tabular}
         }
         \\
         \midrule
         Action &
         \multicolumn{2}{c}{
         \makecell{Fix the inferred exogenous variables, $\epsilon_m$, in the causal graph, \\
         and apply any interventions such that parents of $x_m$, $pa_m$, become $\hat{pa}_m$.}}
         \\
         \midrule
         Prediction &
         $x_{m, \cf} = f_m(\epsilon_m; \hat{pa}_m)$
         &
         \makecell{
         $x_{m, \cf} = l_m(u_m; h_m(z_m, \hat{pa}), \hat{pa})$
         } \\
         \bottomrule
    \end{tabular}
    }
\end{table}

\section{Implementation of Causal Shape Models}

\subsection{Low-Dimensional Covariate Mechanisms}
\label{app:metadata_mechanisms}
$f_B(\cdot)$, $f_V(\cdot)$ and $f_A(\cdot)$ are modelled as invertible, explicit mechanisms, and implemented using normalising flows \cite{papamakarios2021normalizing}. These mechanisms are one-dimensional and their exogenous noise components $\epsilon$ are sampled from a standard normal. Since $a$ is at the root of the causal graph and has no causes, $f_A(\cdot)$ is implemented by an unconditional flow,
\begin{gather}
    a \defeq \mech_A(\epsilon_A) = \big(\exp \circ \text{AffineNormalisation} \circ \text{Spline}_\theta \big)(\epsilon_A),
    \label{eq:a_subflow}
\end{gather}
which can be split into the sub-flows $a = \big(\exp \circ \text{AffineNormalisation}\big) (\hat{a})$ and $\hat{a} = \text{Spline}_{\theta}(\epsilon_A)$, where $\text{Spline}_\theta$ refers to a linear spline flow \cite{dolatabadi2020invertible,durkan2019neural}, and AffineNormalisation is a whitening operation in an unbounded log-space. Since $a \rightarrow b \leftarrow s$, $f_B(\cdot)$ is implemented as a conditional flow,
\begin{align}
    b &\defeq \mech_B(\epsilon_B; s, a) \\
    &= \big({\exp \circ \text{AffineNormalisation} \circ \text{ConditionalAffine}_{\theta}([s, \hat{a}])} \big)(\epsilon_B), \notag
    \label{eq:b_subflow}
\end{align}
which can be split into sub-flows $b = \big(\exp \circ \text{AffineNormalisation}\big)(\hat{b})$ and $\hat{b} = \text{ConditionalAffine}_{\theta}([s, \hat{a}])(\epsilon_B)$. Using $\hat{a}$ instead of $a$ for conditioning improves numerical stability, since the weights in the neural network are much smaller in magnitude than $a \in [40, 70]$. It follows that $f_V(\cdot)$ is implemented as 
\begin{align}
    v &\defeq \mech_V(\epsilon_V; a, b) \\
    &= \big({\exp \circ \text{AffineNormalisation} \circ \text{ConditionalAffine}_{\theta}([\hat{b}, \hat{a}])}\big)(\epsilon_V), \notag
    \label{eq:v_subflow}
\end{align}
which can be split into $v = (\exp \circ \text{AffineNormalisation})(\hat{v})$ and \\ $\hat{v} = \text{ConditionalAffine}_{\theta}([\hat{b}, \hat{a}])(\epsilon_V)$, where $\hat{a}$ and $\hat{b}$ are found by inverting the sub-flows $\hat{a} \rightarrow a$ and $\hat{b} \rightarrow b$. The conditional location and scale parameters for $\mathrm{ConditionalAffine}_{\theta}(\cdot)$ are learned by neural networks with 2 linear layers of 8 and 16 neurons and a LeakyReLU(0.1) activation function. This is the same as the implementation in \cite{pawlowski2020dscm}.

The discrete variable $s$ is also a root in the causal graph, as such $f_S(\cdot)$ need not be invertible. The value of $s$ can be set manually after an abduction step. Its mechanism is
\begin{gather}
    s \defeq \mech_S(\epsilon_S) = \epsilon_S \,,
\end{gather}
where $\epsilon_S \sim \text{Bernoulli}(\theta)$ and $\theta$ is the learned probability of being male.

\subsection{ELBO full proof}
From the causal, graphical model in \cref{fig:graph_ukbb_covariate_sem}, we can state the conditional factorisation of independent mechanisms,
\begin{align}
    p(x, b, v, a, s) &= p(x | b, v) \cdot \underbrace{p(v | a, b) \cdot p(b | s, a) \cdot p(a) \cdot p(s)}_{p(v,b,a,s)}.
\end{align}
The joint including the independent exogenous variable $z_X$ can be factorised as
\begin{align}
    p(x, z_X, b, v, a, s) &= p(x | z_X, b, v) \cdot p(z_X) \cdot p(v,b,a,s),
\end{align}
where $z_X$ can be integrated out as
\begin{align}
    p(x, b, v, a, s) &= \int p(x | z_X, b, v) \cdot p(z_X) \cdot p(v,b,a,s) \; dz_X.
\end{align}
We can now formulate the lower bound on the log-evidence $\log p(x, b, v, a, s)$ which can be written as
\begin{align}
    &\log p(x, b, v, a, s) \notag \\
    &\qquad = \log \int p(x | z_X, b, v) \cdot p(z_X) \cdot p(v,b,a,s) \; dz_X \\
    &\qquad = \alpha + \log \int p(x | z_X, b, v) \cdot p(z_X) \; dz_X,
\end{align}
where $\alpha = \log p(v, b, a, s)$. Since the marginalisation over $z_X$ is intractable, we introduce the variational distribution $q(z_X | x, v, b) \approx p(z_X | x, v, b)$,
\begin{align}
    &\log p(x, b, v, a, s) \notag \\
    &\qquad = \alpha + \log \int p(x | z_X, b, v) \cdot p(z_X) \cdot \frac{q(z_X | x, v, b)}{q(z_X | x, v, b)}\; dz_X \\
    &\qquad = \alpha + \log \mathrm{E}_{q(z_X|x, v, b)} \Bigg[ p(x | z_X, b, v) \cdot \frac{p(z_X)}{q(z_X | x, v, b)} \Bigg],
\end{align}
then by Jensen's inequality,
\begin{align}
    \label{eq:jensens}
    &\qquad \geq \alpha + \mathrm{E}_{q(z_X|x, v, b)} \Bigg[ \log \Bigg(p(x | z_X, b, v) \cdot \frac{p(z_X)}{q(z_X | x, v, b)} \Bigg) \Bigg] \\
    \label{eq:elbo_full}
    &\qquad = \alpha + \mathrm{E}_{q(z_X|x, v, b)} \big[ \log p(x, z_X | b, v) - \log q(z_X | x, v, b) \big],
\end{align}
we arrive at a formulation for the evidence lower bound (ELBO) (\cref{eq:elbo_full}). This is optimised directly using stochastic variational inference (SVI) in Pyro using an Adam optimiser, with gradient estimators constructed using the formalism of stochastic computational graphs \cite{schulman2015gradient}. Following \cref{eq:jensens}, the ELBO can also be written using the Kullback-Leibler (KL) divergence,
\begin{align}
    &\qquad = \alpha + \mathrm{E}_{q(z_X|x, v, b)} [\log p(x | z_X, b, v)] - \mathrm{E}_{q(z_X|x, v, b)} \Bigg[ \log \frac{q(z_X | x, v, b)}{p(z_X)} \Bigg] \\
    &\qquad = \alpha + \mathrm{E}_{q(z_X|x,b,v)}[\log p(x | z_X, b, v)] - \mathrm{KL}[q(z_X | x, b, v) \| p(z_X)] \\
    &\qquad = \alpha + \underbrace{{\mathrm{E}_{q(z_X|x,b,v)}[\log p(x | z_X, b, v)] - \mathrm{KL}[q(z_X | x, b, v) \| \mathcal{N}(0, I_{D})]}}_{\beta},
\end{align}
which clearly demonstrates that $\beta$ learns a mesh CVAE within the CSM structure. The density $p(x|z_X, b, v)$ is computed using the change of variables rule as
\begin{align}
    p(x | z_X, b, v) = p(u_X) \cdot |\text{det} \: \nabla_{u_X} l_X(u_X; \conddec(z_X; v, b))|\inv,
\end{align}
with $u_X = l_X\inv(x; \: \conddec(z_X; v, b))$ 

\section{Training Causal Shape Models}
\label{app:implementation}

\begin{figure}[h]
    \begin{center}
        \includegraphics[width=.6\textwidth]{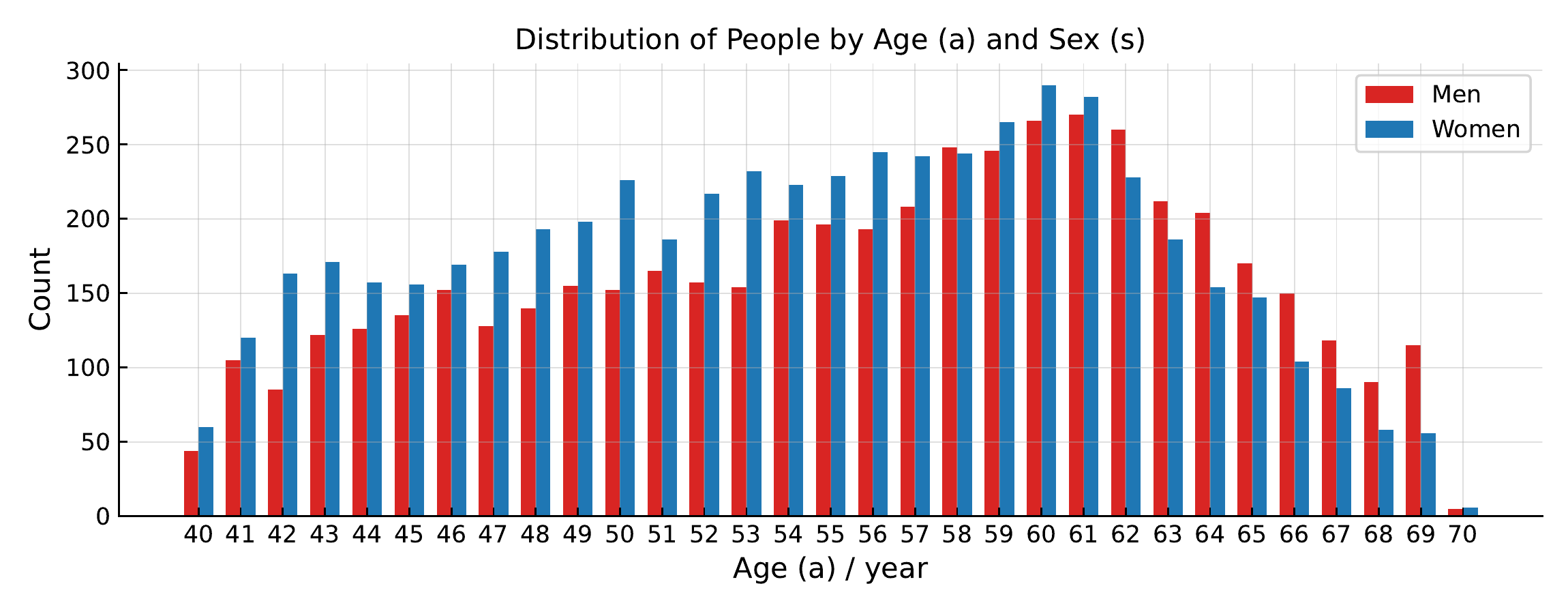}
        \caption{Distribution of individuals by age and sex in our training subset of the UK Biobank Imaging Study.}
        \label{fig:true_distribution}
    \end{center}
\end{figure}

\paragraph{\bf Dataset.} Our dataset consists of the age ($a$), biological sex ($s$), total brain volume ($b$), brain stem volume ($v$) and the corresponding triangulated brain stem and 4th ventricle surface mesh ($x$) (simply referred to as a \textit{brain stem}) for 14,502 individuals from the UK Biobank Imaging Study \cite{sudlow2015uk}. The brain structures had been automatically segmented from the corresponding brain MRI scans. Each brain stem mesh consists of 642 vertices, and we randomly divide the dataset into 10,441/1,160/2,901 meshes corresponding to train/validation/test splits. In the training set, there are 5471 females and 4970 males. The true age and sex distributions are presented in \cref{fig:true_distribution}. The results in this paper are produced using the test set.

\paragraph{\bf Training Considerations.} The $\text{ChebConv}_{\theta}(\cdot)$ demonstrates sensitivity to vertex degrees, due to the underlying graph Laplacian operator \cite{zhang2010spectral}, resulting in spikes at irregular vertices in the early stages of training. We noticed 12 spikes corresponding to the 12/642 vertices that are 5-regular, with the remaining 6-regular vertices being reconstructed smoothly. In our case, it sufficed to tune the learning rate and $K$ carefully, given that our mesh's topology was already very smooth. \cite{nicolet2021large} and \cite{yuan2020mesh} proposed a number of solutions to this problem which should be considered in future work, if we continue to use spectral graph convolutions.

\section{Further Experiments with Brain Stem Meshes}
We use Scikit-learn's implementation of PCA \cite{scikit-learn} throughout. The vertex Euclidean distance (VED) between meshes $x, x' \in \mathbb{R}^{|V| \times 3}$ is defined as
\begin{align}
    \text{VED}(x, x') = \frac{\sum^{|V|}_{v=1}\sqrt{(x_{v,1} - x'_{v,1})^2 + (x_{v,2} - x'_{v,2})^2 + (x_{v,3} - x'_{v,3})^2}}{|V|}
\end{align}

{\label{app:experiments}}

\clearpage
\subsection{Association}

 \begin{figure}[h!]
    \begin{center}
        \begin{subfigure}{\textwidth}
            \makebox[\textwidth]{
                \includegraphics[width=0.8\textwidth]{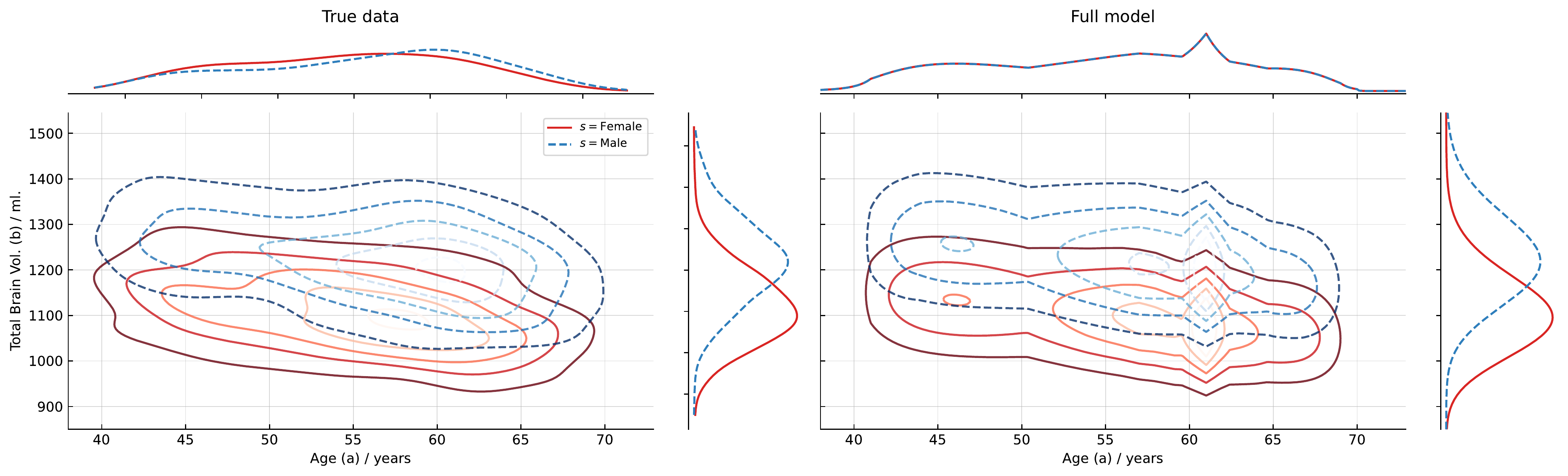}
            }
            \subcaption{Age ($a$) vs.\ total brain volume ($b$): $p(a, b\,|\, s)$. Here we see differences in brain volume across biological sexes, as well as a downward trend in brain volume as age progresses.}
            \label{fig:medical_age_bvol}
        \end{subfigure}
    \end{center}
    \begin{center}
    \begin{subfigure}{\textwidth}
            \makebox[\textwidth]{
                \includegraphics[width=0.8\textwidth]{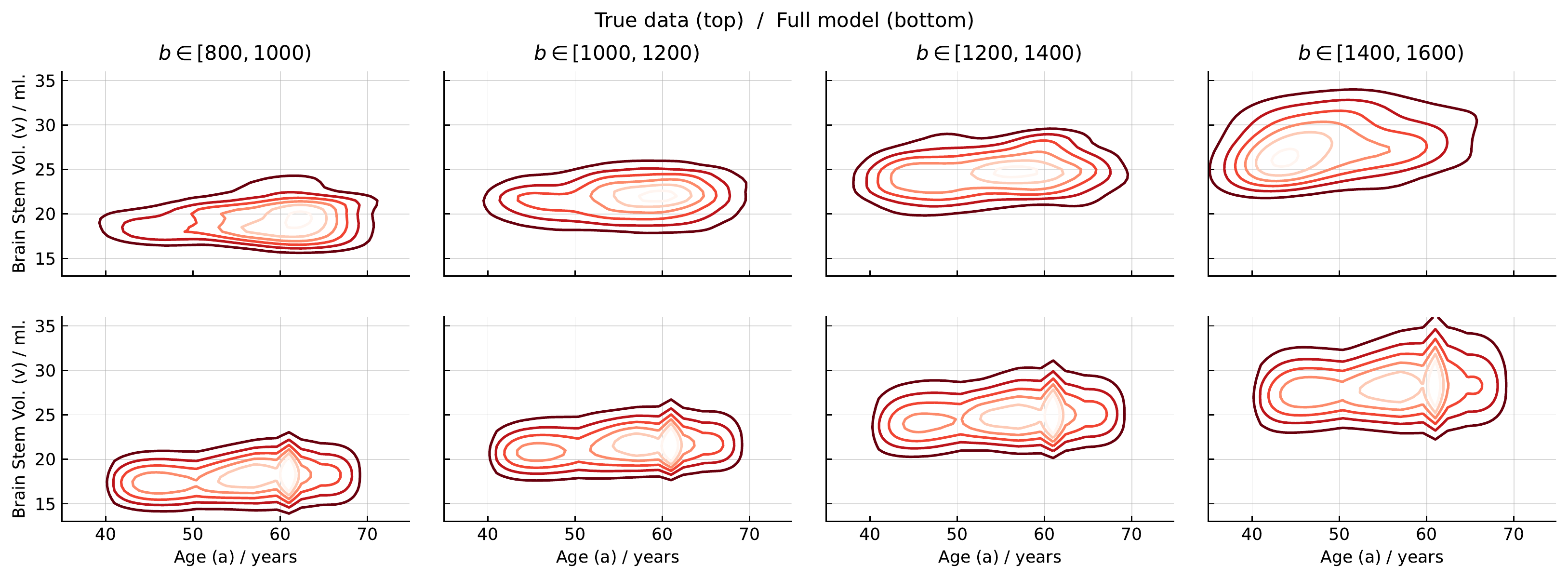}
            }
        \subcaption{Age ($a$) vs.\ brain stem volume ($v$): $p(a, v \,|\, b \in \cdot\,)$. We observe a consistent increase in brain stem volume with age, in addition to a proportionality relationship with the overall brain volume.}
        \label{fig:medical_age_vvol}
    \end{subfigure}
    \end{center}
    \begin{center}
    \begin{subfigure}{\textwidth}
            \makebox[\textwidth]{
                \includegraphics[width=0.8\textwidth]{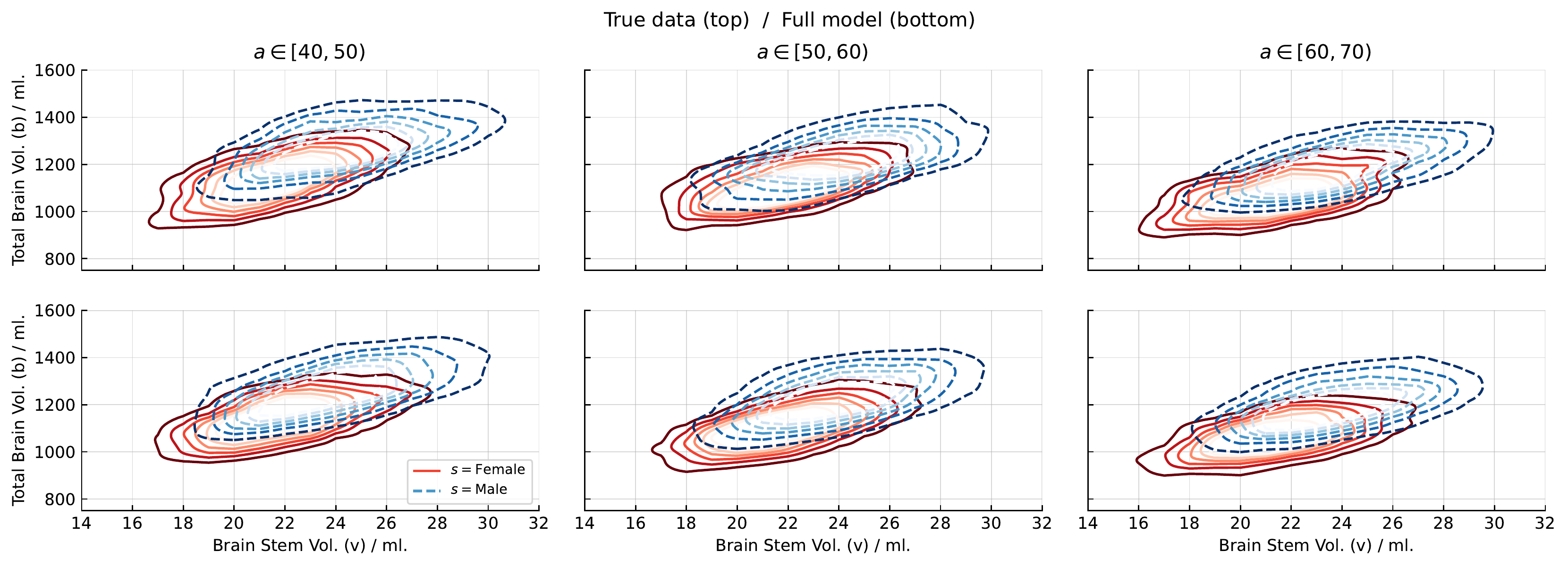}
            }
        \subcaption{Total brain volume (b) vs brain stem volume (v): $p(b, v\,|\, a \in \cdot\, , s)$. Here we see the positive correlation between brain stem volume and total brain volume, which is the present across all age and sex sub-populations.}
        \label{fig:medical_age_vvol_2}
    \end{subfigure}
    \end{center}
\caption{Diagrams and captions taken and adapted with permission from \cite{pawlowski2020dscm}. Densities for the true data (KDE) and for the learned model. The overall trends and interactions present in the true data distribution seem faithfully captured by the model.}
\label{fig:learned_distributions}
\end{figure}

\begin{table*}[h]
    \centering
    \caption{Reconstruction errors between $x$ and $\tilde{x}$ for PCA models and $f_X(\cdot)$. All values are in millimetres (mm).}
    \setlength{\tabcolsep}{.5em}
    \scalebox{0.80}{
        \begin{tabular}{lcccc}
        \toprule
        Model Type & Latent Dim. & Mean VED $\pm$ Error & Median VED & Chamfer Distance \\
        \midrule
        PCA                  & 8  & 0.5619 $\pm$ 0.12 & 0.5473 & 0.8403 \\
                             & 16 & 0.3951 $\pm$ 0.08 & 0.3862 & 0.4252 \\
                             & 32 & 0.1827 $\pm$ 0.05 & 0.1773 & 0.0903 \\
                             & 64 & 0.0194 $\pm$ 0.01 & 0.0176 & 0.0012 \\ 
        \midrule
        $f_X(z_X^i, u_X; \cdot)$        & 8  & 2.3834 $\pm$ 0.15 & 2.3921 & 11.4013 \\
                                        & 16 & 0.9006 $\pm$ 0.10 & 0.8900 & 2.0421 \\
                                        & 32 & 1.3844 $\pm$ 0.09 & 1.3766 & 4.1885 \\
                                        & 64 & 3.5714 $\pm$ 0.36 & 3.5783 & 16.1302 \\
        \midrule
        $f_X(z_X^i, u_X^i; \cdot)$ &  8, 16, 32, 64  & 0 & 0 & 0 \\
        \bottomrule
        \end{tabular}
    }
    \label{table:reconstruction_results}
\end{table*}

\begin{figure}[h]
    \begin{center}
        \includegraphics[width=.8\textwidth]{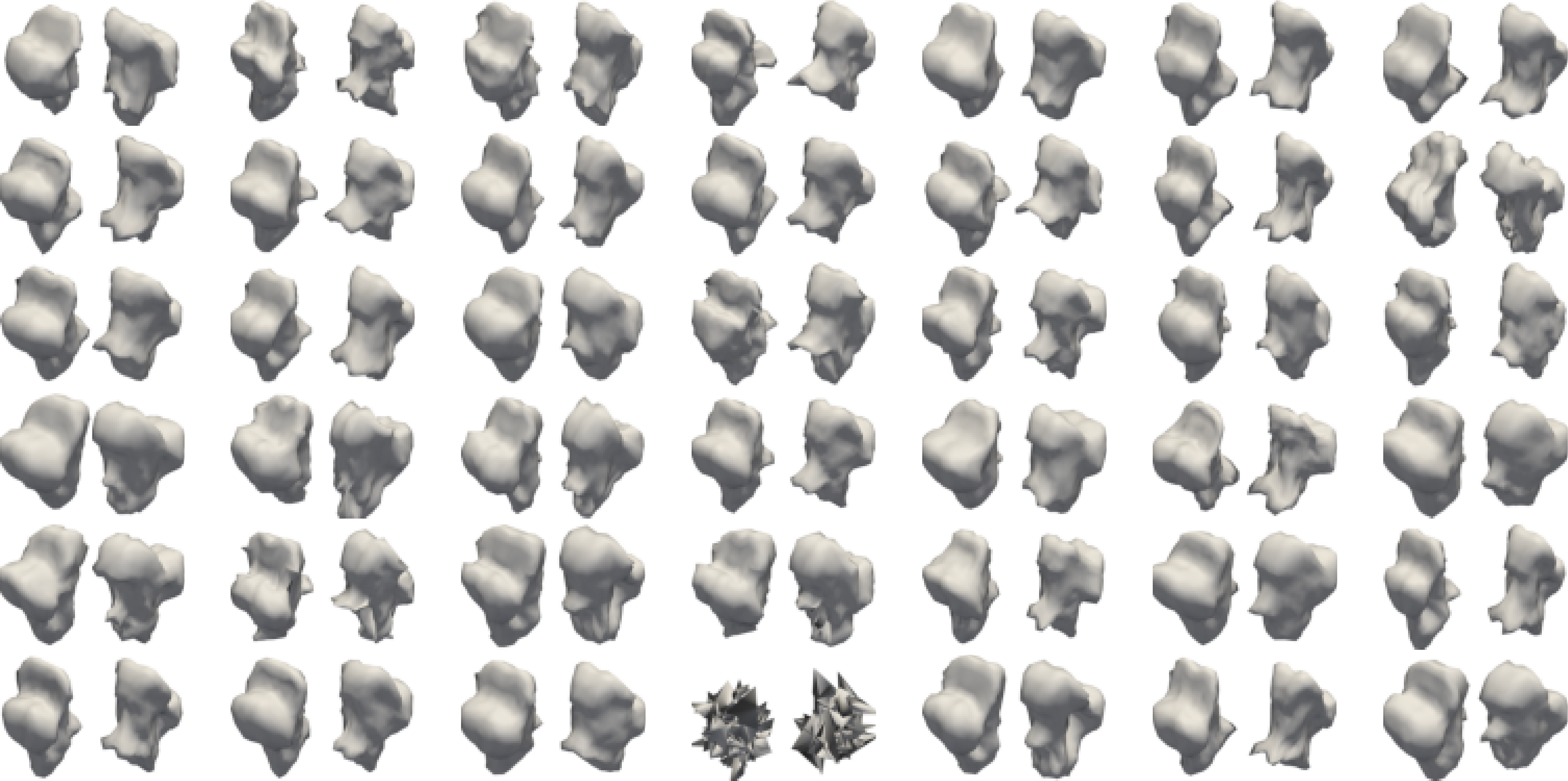}
        \caption{Random samples from the CSM. Each column includes the anterior (left) and posterior (right) view. In the middle column on the bottom row, the decoder in the CSM was unable to generalise the sampled $z_X$, resulting in an anomalous brain stem shape being generated.}
        \label{fig:random_samples}
    \end{center}
\end{figure}

\clearpage
\subsection{Population-Level}

\begin{figure}[h]
\centering
\begin{minipage}{.5\textwidth}
    \centering
    \includegraphics[width=\textwidth]{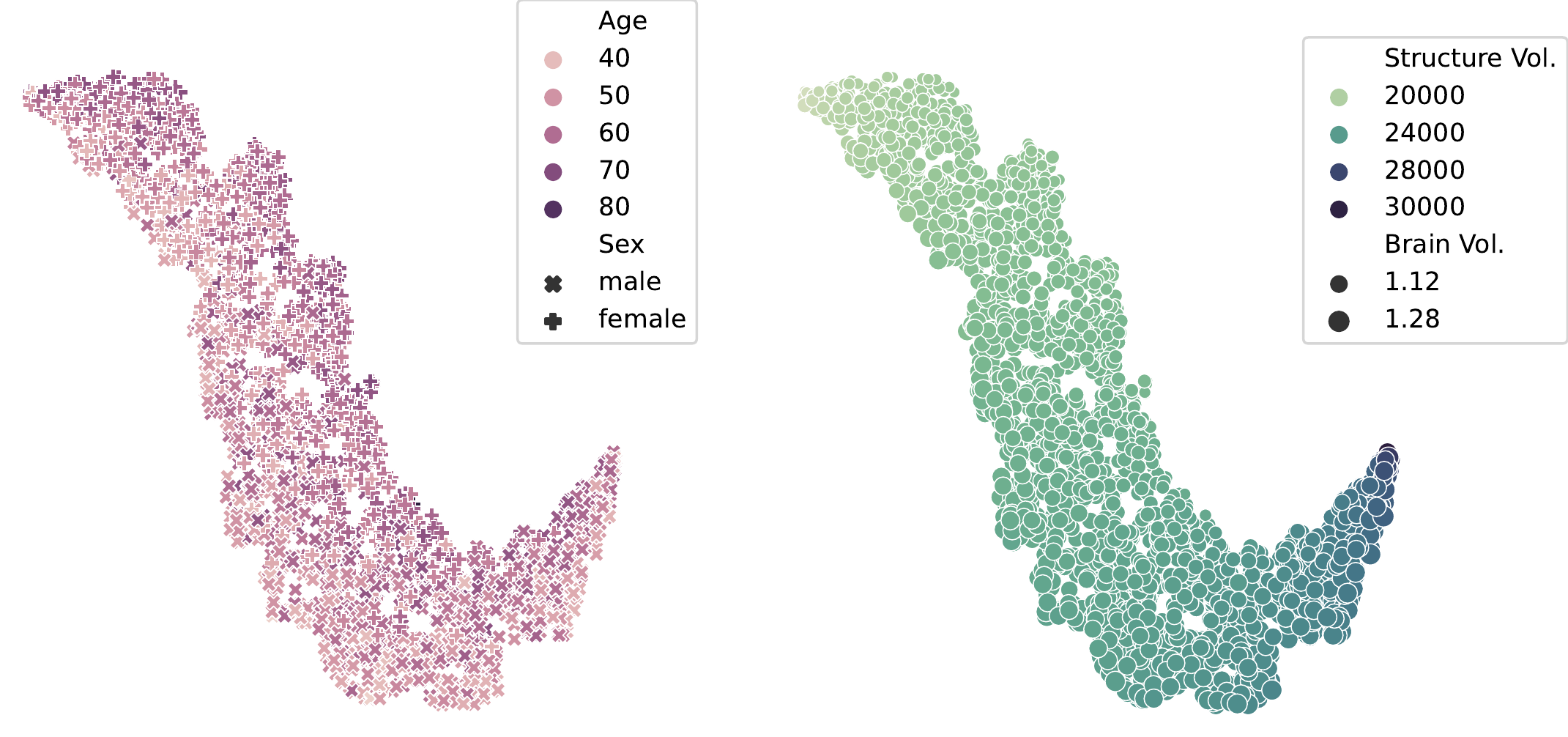}
    \captionof{figure}{t-SNE projection following the procedure used for \cref{fig:shape_projection_population_level} using a different $z_X$.}
    \label{fig:shape_projection_population_level_2}
\end{minipage}
\begin{minipage}{.4\textwidth}
    \centering
    \captionof{table}{Average specificity errors following the procedure used for \cref{tab:specificity} using $z_X$ from \cref{fig:shape_projection_population_level_2}.}
    \label{tab:specificity_2}
    \begin{tabular}{lcc}
        \toprule
        $do(\cdot)$ & Mean $\pm$ Error & Median \\
        \midrule
        $do(a, s)$    & 1.339 $\pm$ 0.275 & 1.239 \\
        $do(b, v)$    & 3.037 $\pm$ 1.503 & 2.839 \\
        \bottomrule
    \end{tabular}
\end{minipage}
\label{fig:tsne2}
\end{figure}
\begin{figure}[h]
\centering
\begin{minipage}{.5\textwidth}
    \centering
    \includegraphics[width=\textwidth]{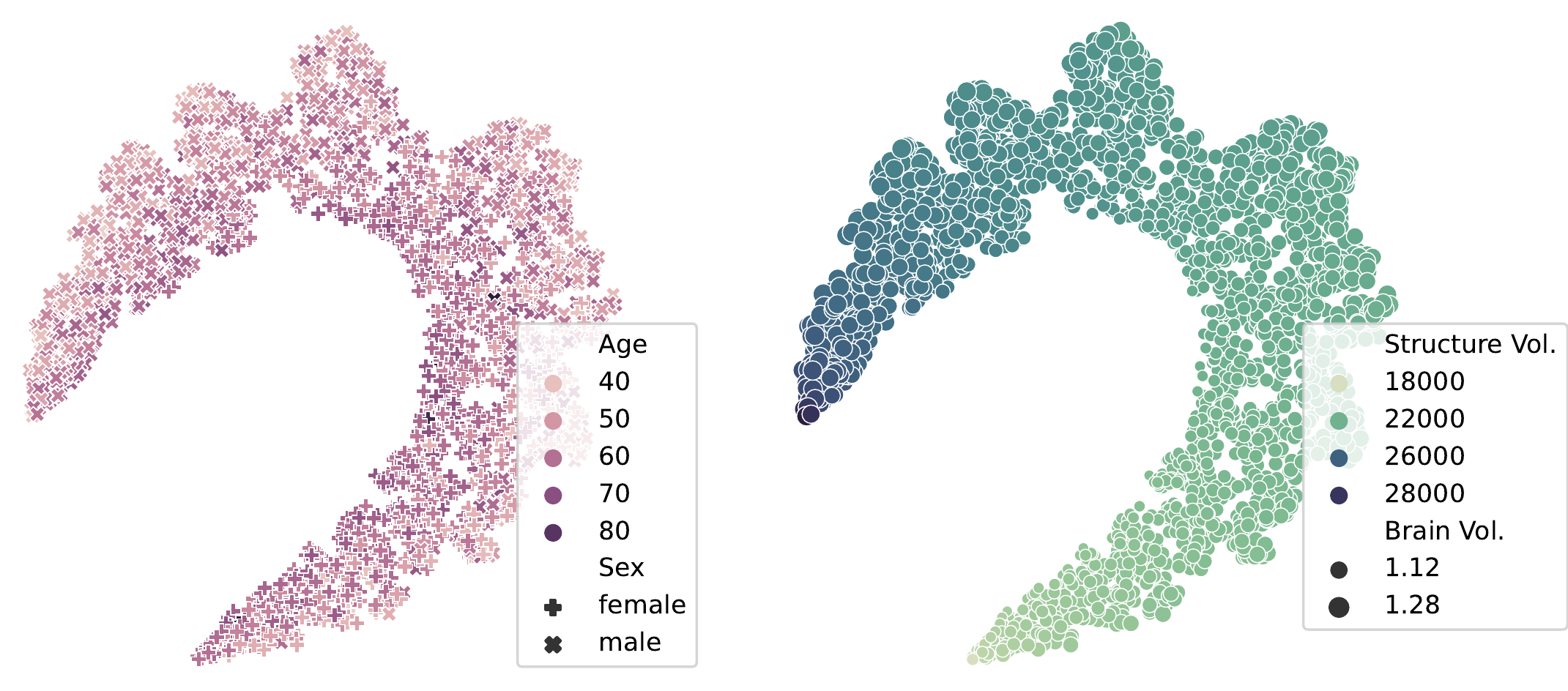}
    \captionof{figure}{t-SNE projection following the procedure used for \cref{fig:shape_projection_population_level} using a different $z_X$.}
    \label{fig:shape_projection_population_level_3}
\end{minipage}
\begin{minipage}{.4\textwidth}
    \centering
    \captionof{table}{Average specificity errors following the procedure used for \cref{tab:specificity} using $z_X$ from \cref{fig:shape_projection_population_level_3}.}
    \label{tab:specificity3}
    \begin{tabular}{lcc}
        \toprule
        $do(\cdot)$ & Mean $\pm$ Error & Median \\
        \midrule
        $do(a, s)$    & 1.261 $\pm$ 0.177 & 1.200 \\
        $do(b, v)$    & 2.238 $\pm$ 0.872 & 2.054 \\
        \bottomrule
    \end{tabular}
\end{minipage}
\label{fig:tsne3}
\end{figure}

In \cref{sec:interventional_results}, we demonstrate that the full range of interventions produce realistic meshes for a chosen $z_X \approx 0$. We see similar trends in \cref{fig:shape_projection_population_level_2} and \cref{fig:shape_projection_population_level_3} for other mesh shapes, $z_X$, which $\conddec(\cdot)$ can generalise to.

Due to the assumed causal, graphical model, our CSM need not produce realistic meshes from the $A$ and $S$ sub-populations under interventions $do(s=S)$ and $do(a=A)$. Generated meshes nevertheless present shapes associated with $A$ and $S$ settings, primarily around the pons, medulla and 4th ventricle, due to learned correlations between volumes ($b$ and $v$) and sub-population specific shapes. This is further explained by \cref{fig:trajectories}, where $do(s=S)$ and $do(a=A)$ result in $v$ and $b$ values from the $A$ and $S$ sub-populations, which in turn deform the observed mesh.

To generate realistic meshes for an age or sex, our CSM would need to include the dependencies $s \rightarrow x$ and $a \rightarrow x$. As a result, our implementation may require architectural changes, such as conditioning each layer in the decoder on the latent style features \cite{regateiro20213d,ma2020learning}, or using a critic at the decoder output \cite{he2019attgan}, akin to the VAE-GAN framework \cite{larsen2016autoencoding}. This will be particularly important when modelling meshes with large deformations, e.g. craniofaical, hand, full body, objects, where one would expect a population-level intervention to generate realistic meshes from a sub-population.

\begin{figure}[h]
    \begin{center}
        \includegraphics[width=.7\textwidth]{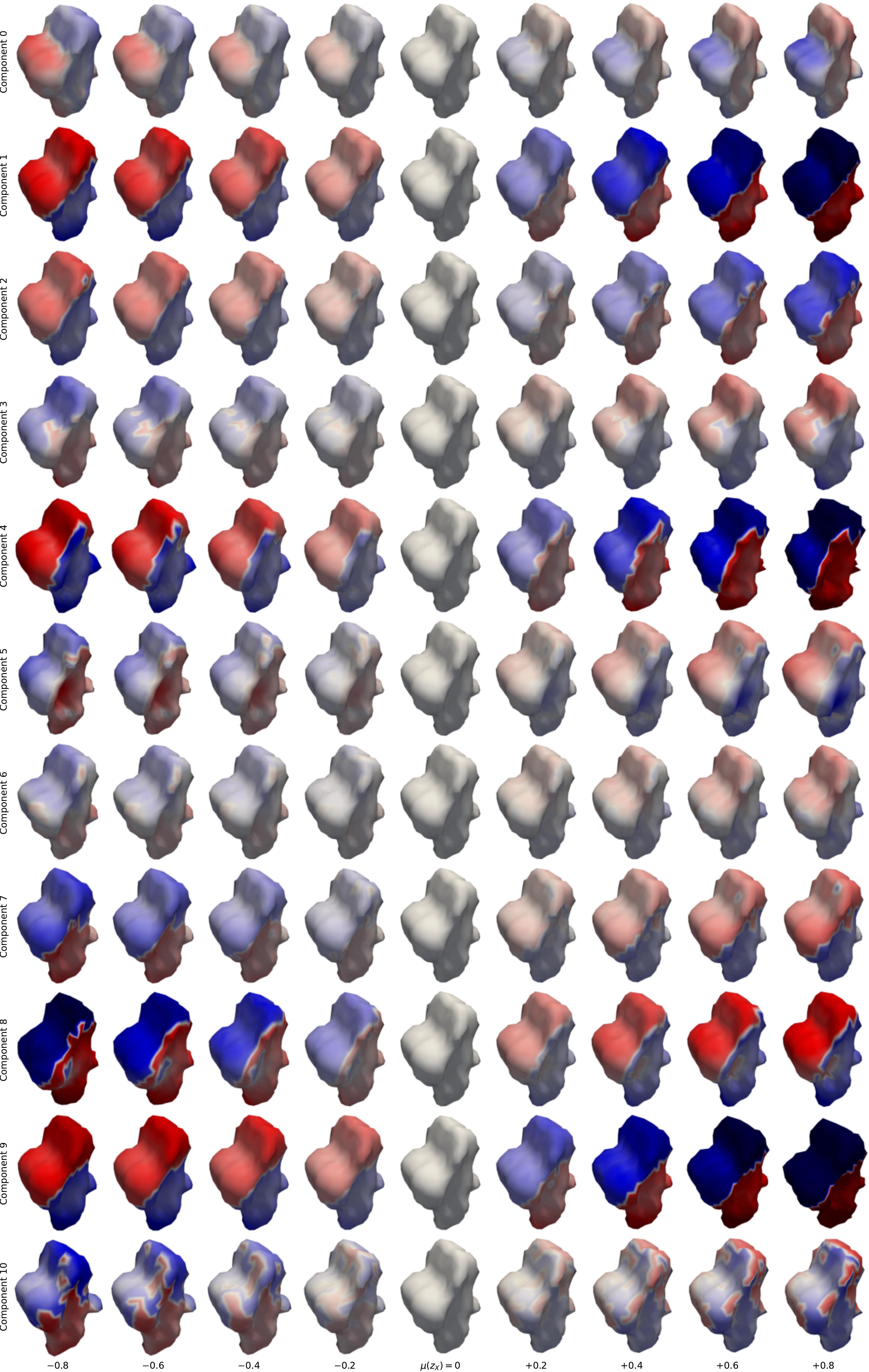}
        \caption{Anterior view of interpolating the latent space $z_X$ following the procedure used for \cref{fig:shape_interpolation_population_level_simplified}.}
    \end{center}
\end{figure}

\begin{figure}[h]
    \begin{center}
        \includegraphics[width=0.7\textwidth]{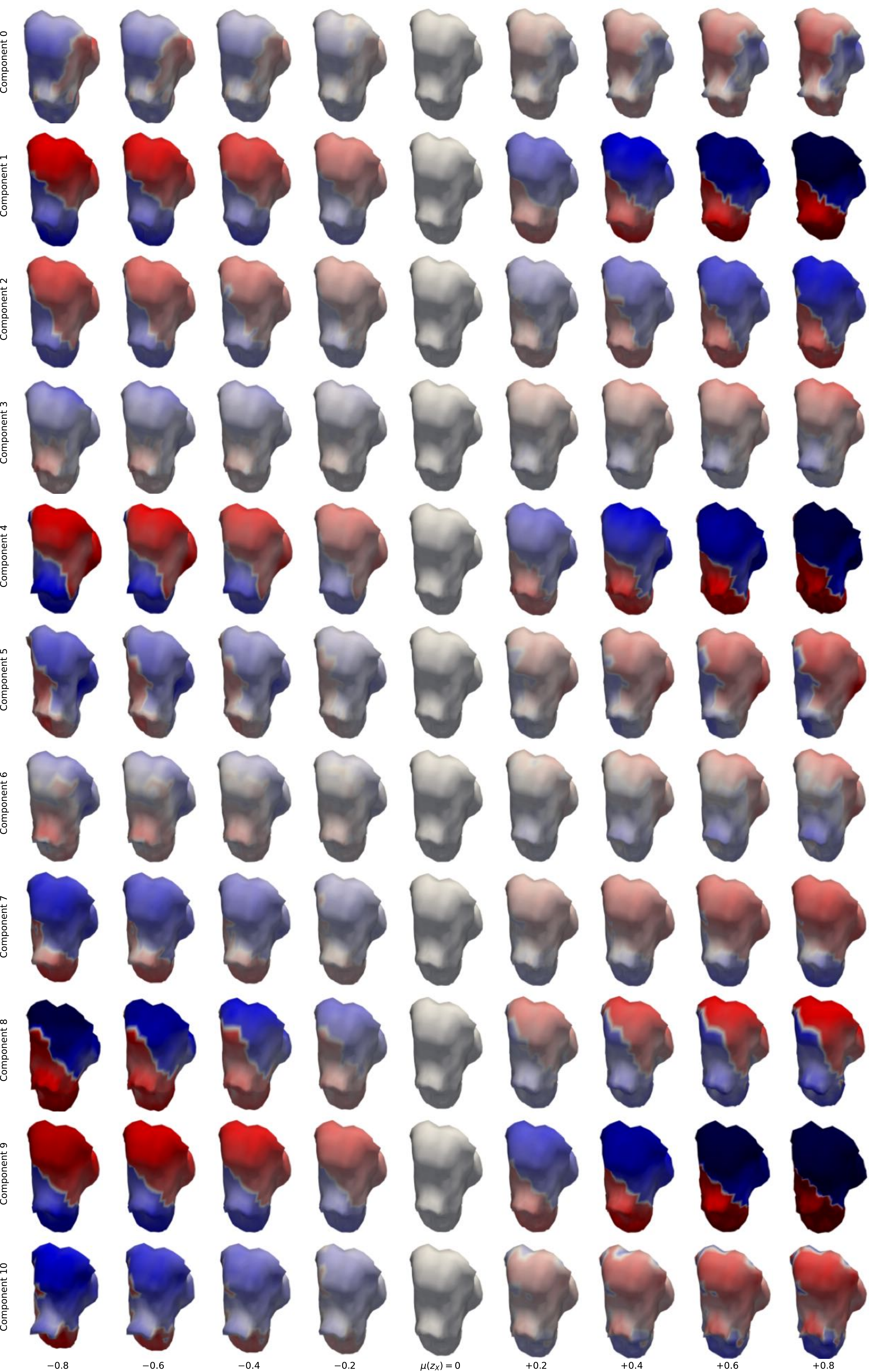}
        \caption{Posterior and lateral view of interpolating the latent space $z_X$ following the procedure used for \cref{fig:shape_interpolation_population_level_simplified}.}
        \label{fig:shape_interpolation_population_level}
    \end{center}
\end{figure}

\clearpage

\subsection{Subject-Specific}

 \begin{figure}[h!]
    \begin{center}
        \begin{subfigure}{\textwidth}
            \makebox[\textwidth]{
                \includegraphics[width=0.7\textwidth]{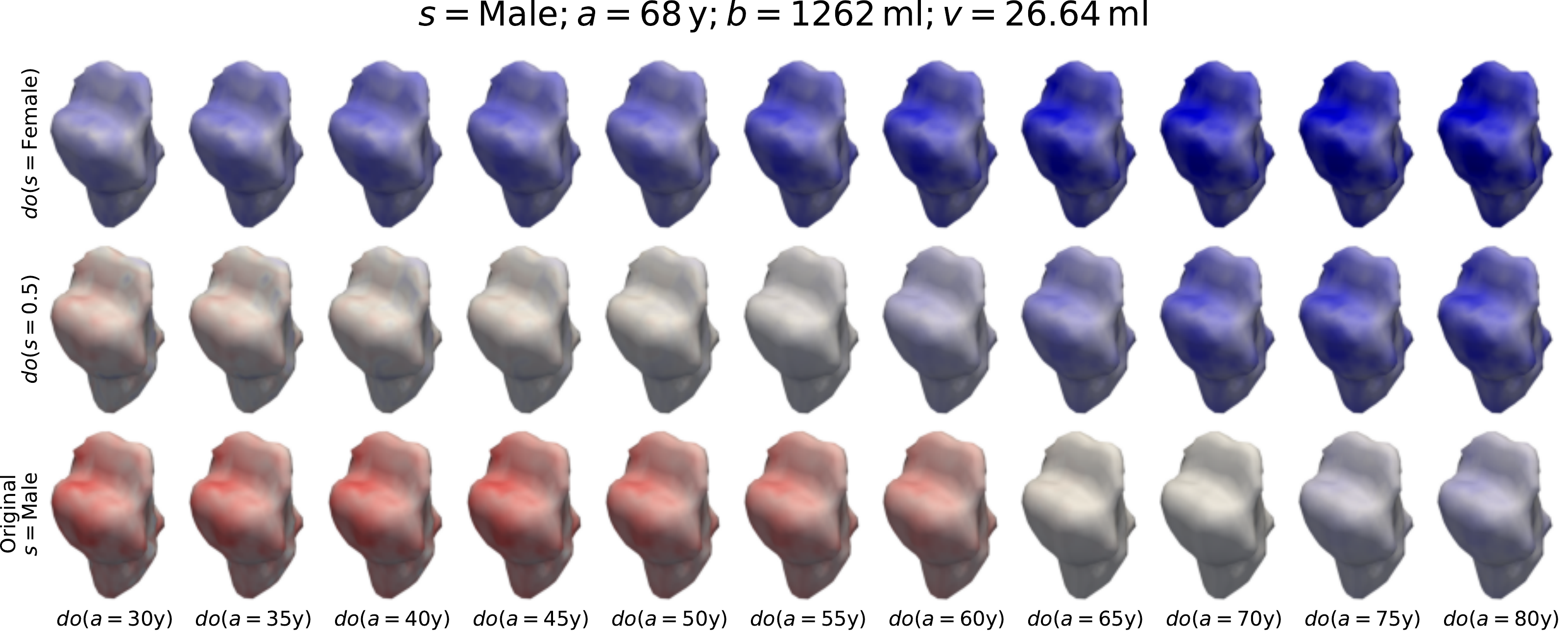}
            }
            \par\bigskip
            \makebox[\textwidth]{
                \includegraphics[width=0.7\textwidth]{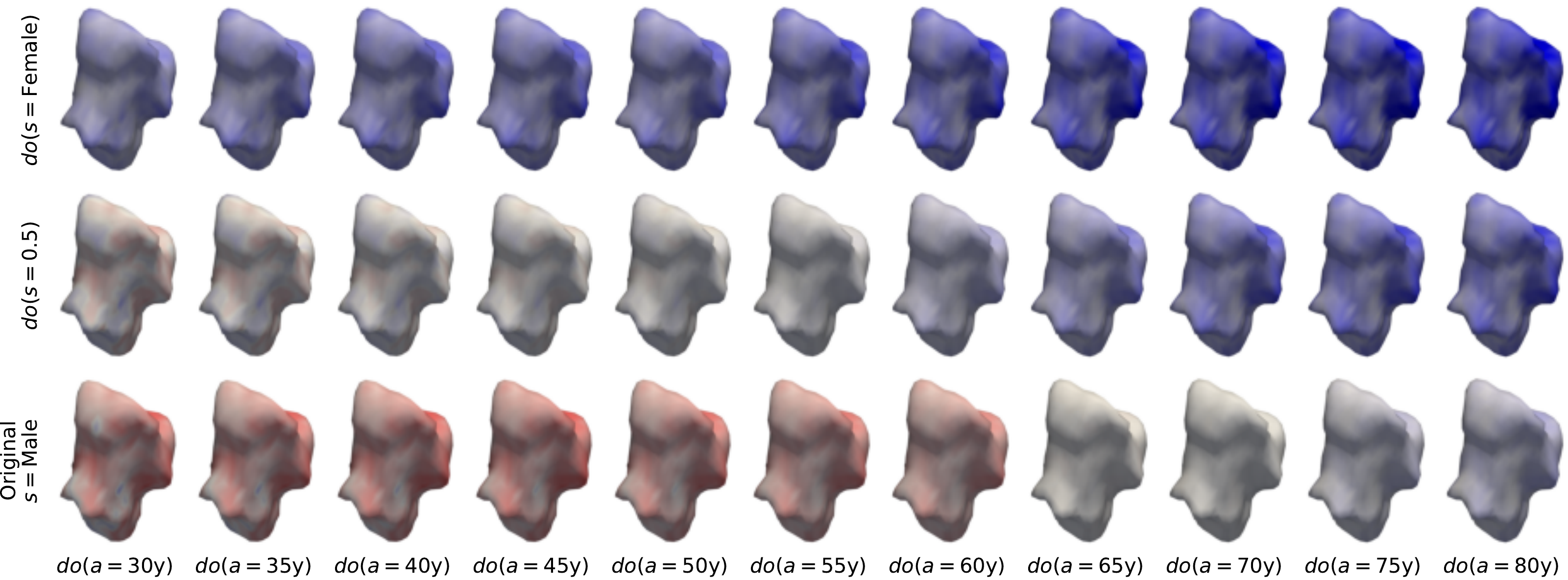}
            }
        \subcaption{Counterfactual meshes for an individual under $do(a)$ and $do(s)$ -- \textit{``What would this person's brain stem look like if they were older/younger or female?"}. \textbf{Top:} Anterior view. \textbf{Bottom:} Posterior and lateral view.}
    \end{subfigure}
    \end{center}
    \begin{center}
        \begin{subfigure}{\textwidth}
            \makebox[\textwidth]{
                \includegraphics[width=0.7\textwidth]{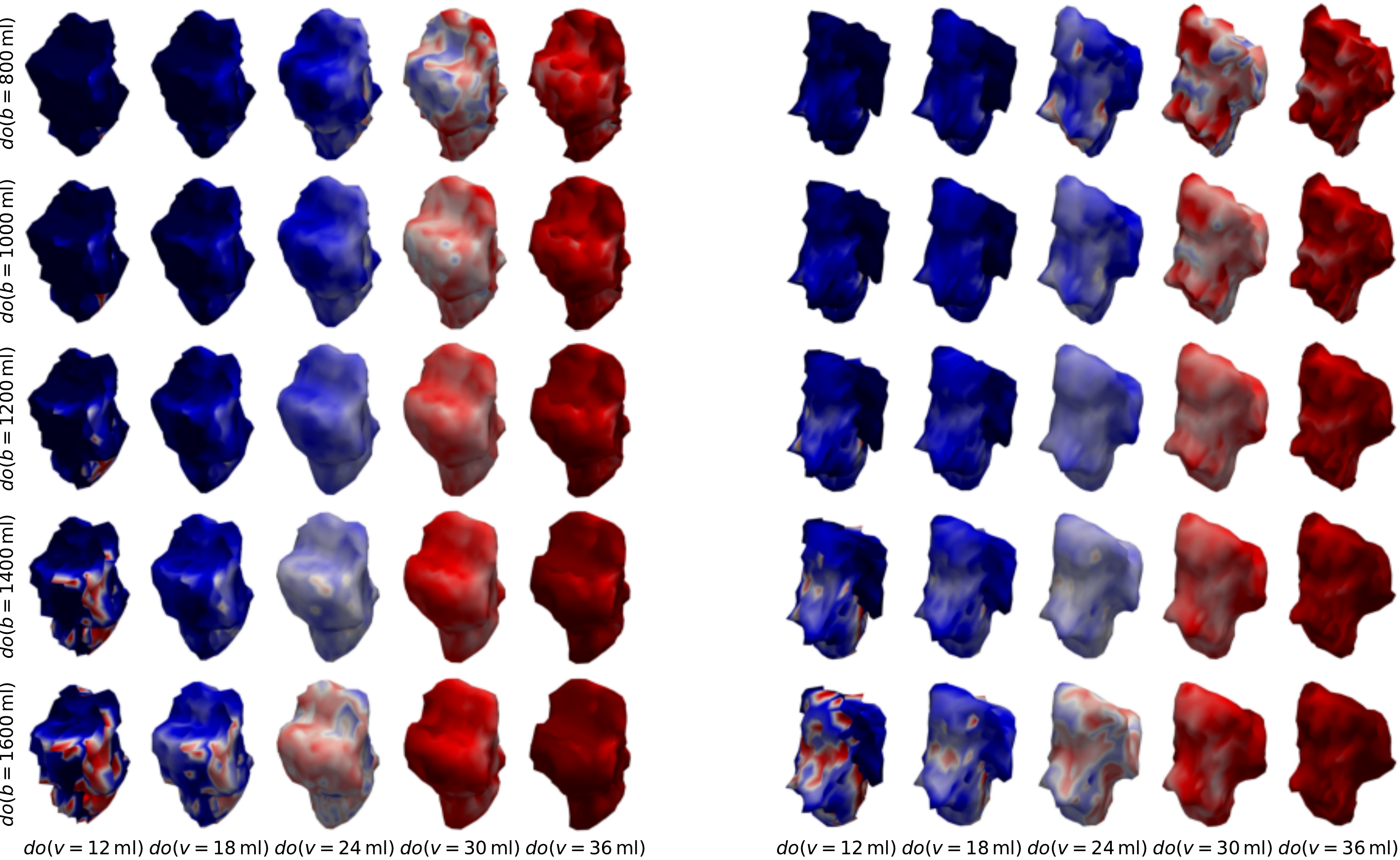}
            }
            \subcaption{Counterfactual meshes for an individual under $do(v)$ and $do(b)$ -- \textit{``What would this person's brain stem look like if the total brain or brain stem volumes were larger or smaller?"}.  \textbf{Left:} Anterior view. \textbf{Right:} Posterior and lateral view.}
        \end{subfigure}
    \end{center}
\caption{\small Counterfactual brain stem meshes for a 68 year old, male. Colours show the VED between observed and counterfactual meshes -- \protect\markertwo\, = +5mm to \protect\markerone\, = -5mm.}
\label{fig:stem_cfs_full_1}
\end{figure}

 \begin{figure}[h!]
    \begin{center}
        \begin{subfigure}{\textwidth}
            \makebox[\textwidth]{
                \includegraphics[width=0.7\textwidth]{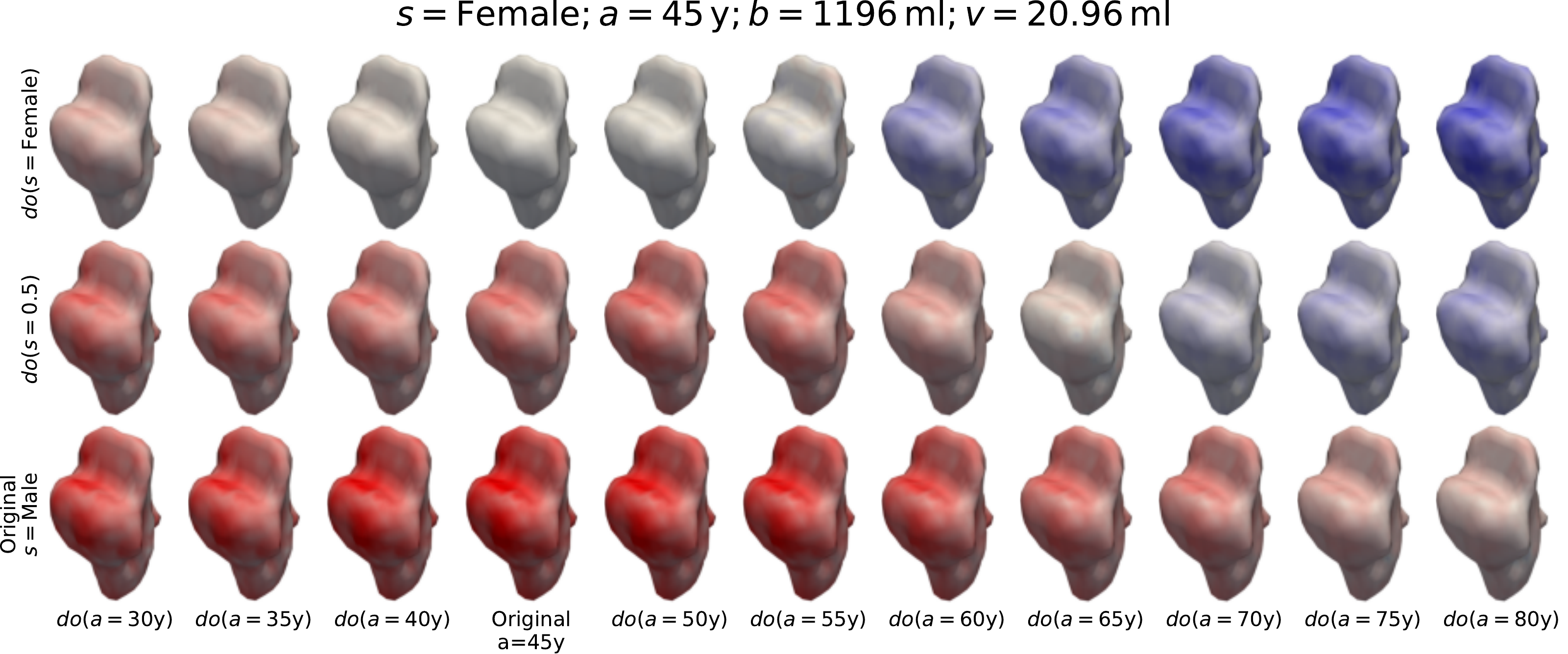}
            }
            \par\bigskip
            \makebox[\textwidth]{
                \includegraphics[width=0.7\textwidth]{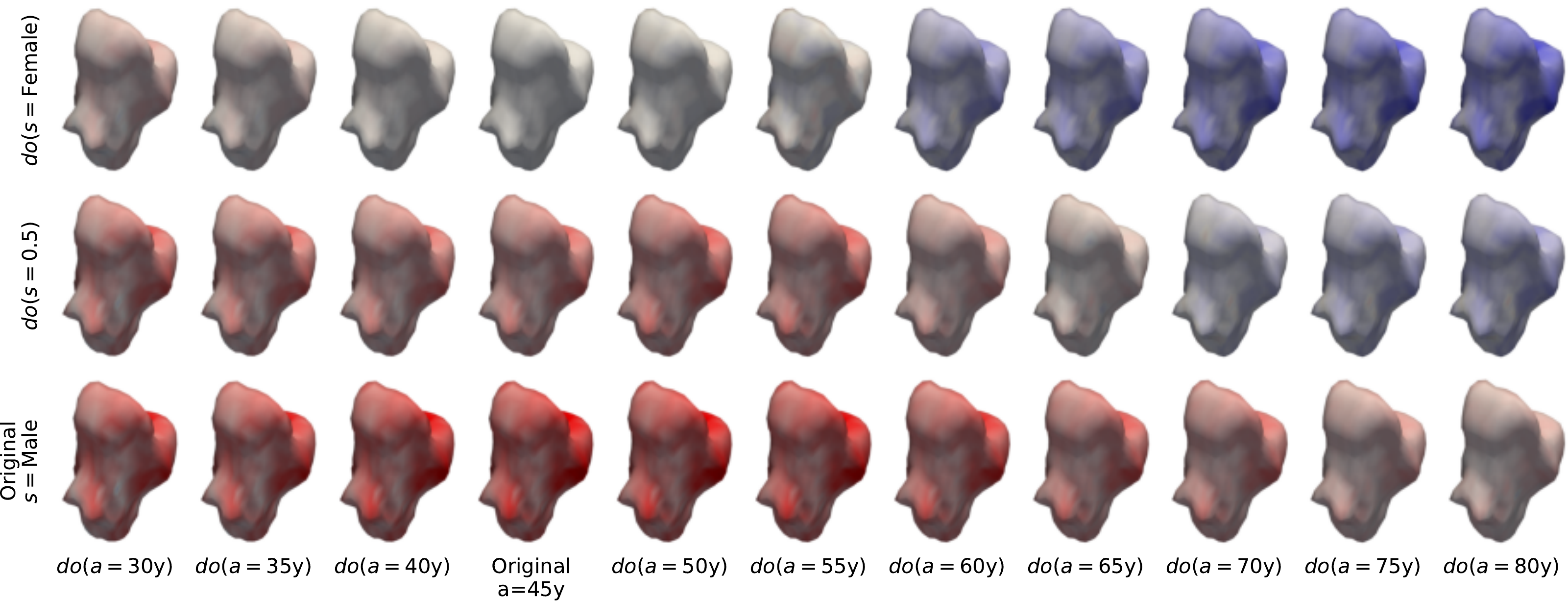}
            }
        \subcaption{Counterfactual meshes for an individual under $do(a)$ and $do(s)$ -- \textit{``What would this person's brain stem look like if they were older/younger or female?"}. \textbf{Top:} Anterior view. \textbf{Bottom:} Posterior and lateral view.}
    \end{subfigure}
    \end{center}
    \begin{center}
        \begin{subfigure}{\textwidth}
            \makebox[\textwidth]{
                \includegraphics[width=0.7\textwidth]{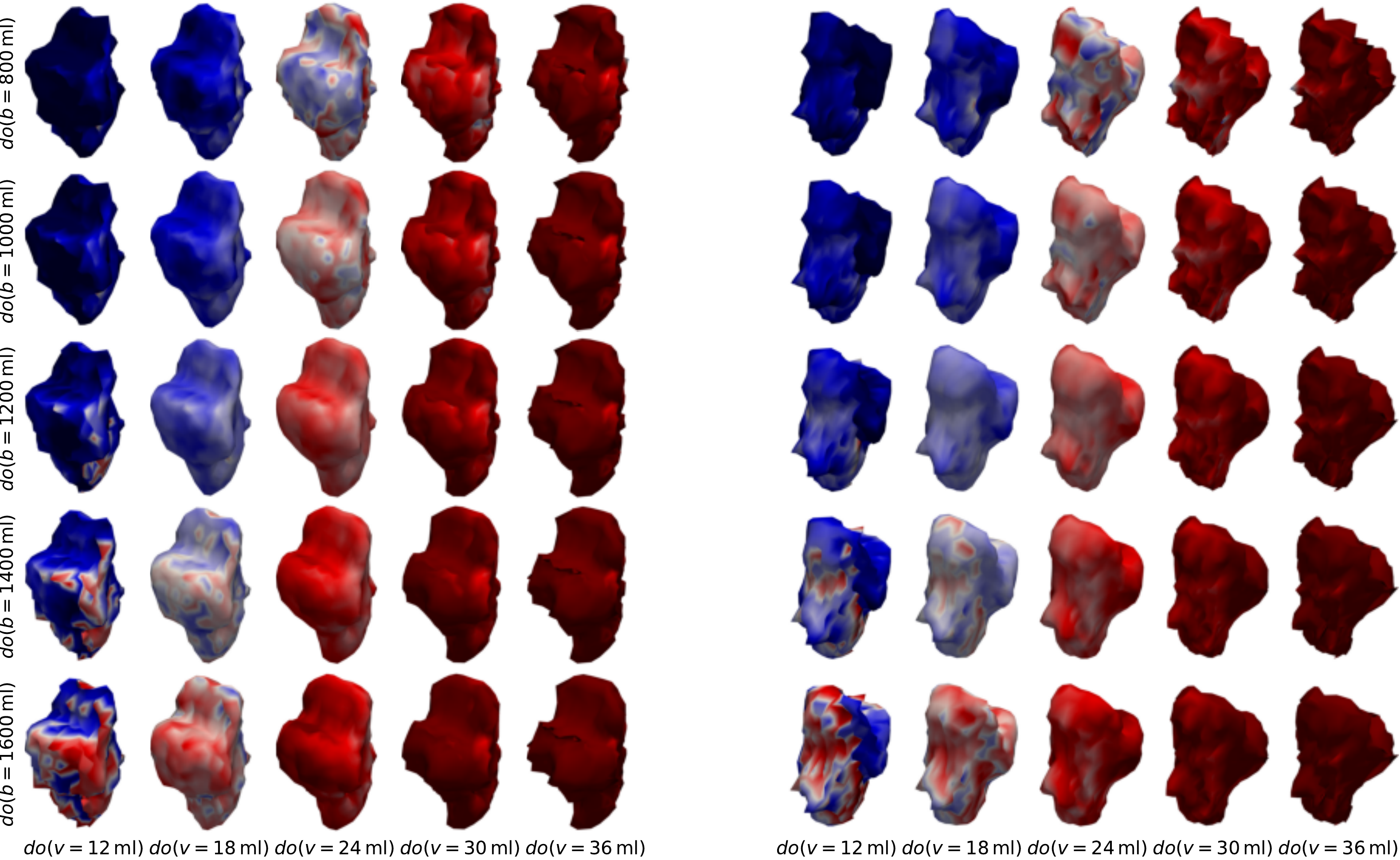}
            }
            \subcaption{Counterfactual meshes for an individual under $do(v)$ and $do(b)$ -- \textit{``What would this person's brain stem look like if the total brain or brain stem volumes were larger or smaller?"}. \textbf{Left:} Anterior view. \textbf{Right:} Posterior and lateral view.}
        \end{subfigure}
    \end{center}
\caption{Counterfactual brain stem meshes for a 45 year old, female. Colours show the VED between observed and counterfactual meshes -- \protect\markertwo\, = +5mm to \protect\markerone\, = -5mm.}
\label{fig:stem_cfs_full_2}
\end{figure}

\clearpage

\section{Preliminary Results on Hippocampus Meshes}
In this section, we present preliminary, qualitative results for a CSM of right hippocampus meshes (simply referred to as the \textit{hippocampus}), learned using the same set of individuals from the UK Biobank Imaging Study. We assume the causal graph in \cref{fig:graph_ukbb_covariate_sem}, where $v$ is now the volume of the hippocampus and $x$ is the hippocampus mesh. We adapt the architecture in \cref{sec:coma} to account for hippocampus meshes having 664 vertices, $|V| = 664$. The diagrams in this section are produced by a CSM with $D = 64$.

\begin{figure}[h]
    \begin{center}
        \begin{subfigure}{\textwidth}
            \makebox[\textwidth]{
                \includegraphics[width=0.7\textwidth]{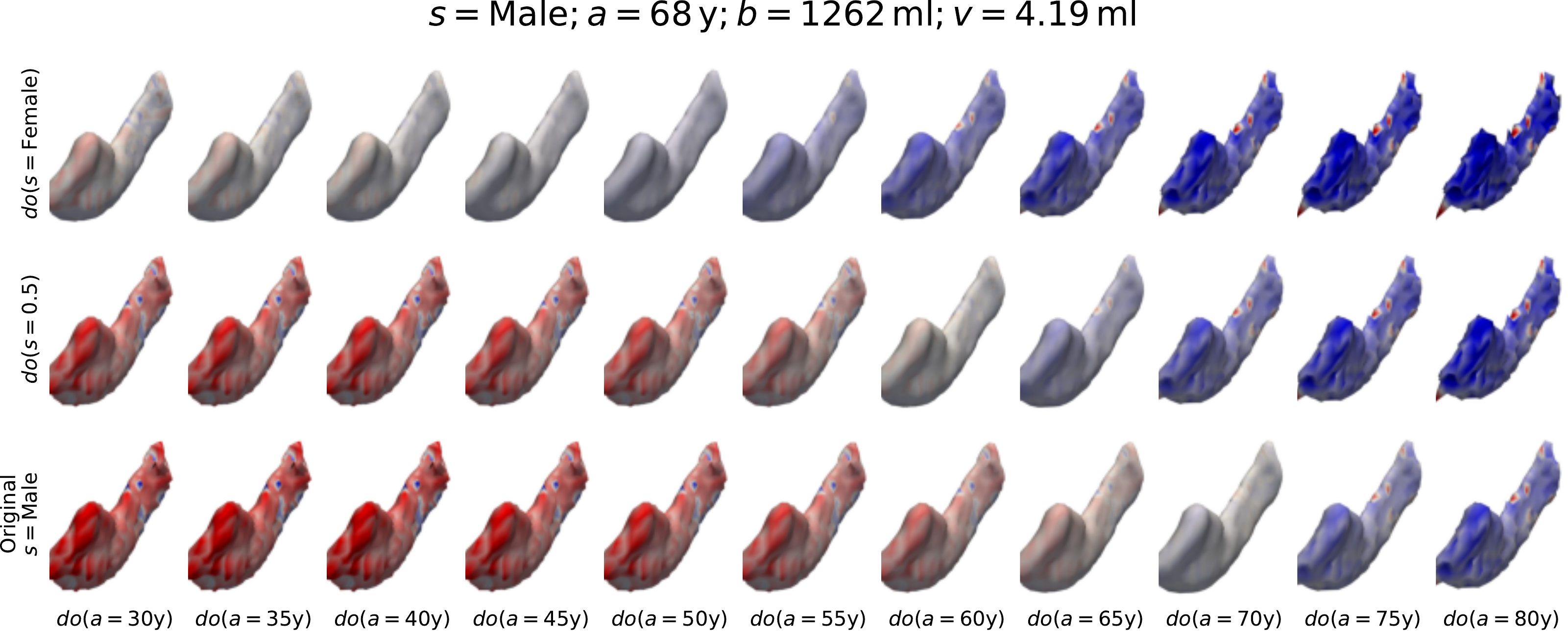}
            }
            \par\bigskip
            \makebox[\textwidth]{
                \includegraphics[width=0.7\textwidth]{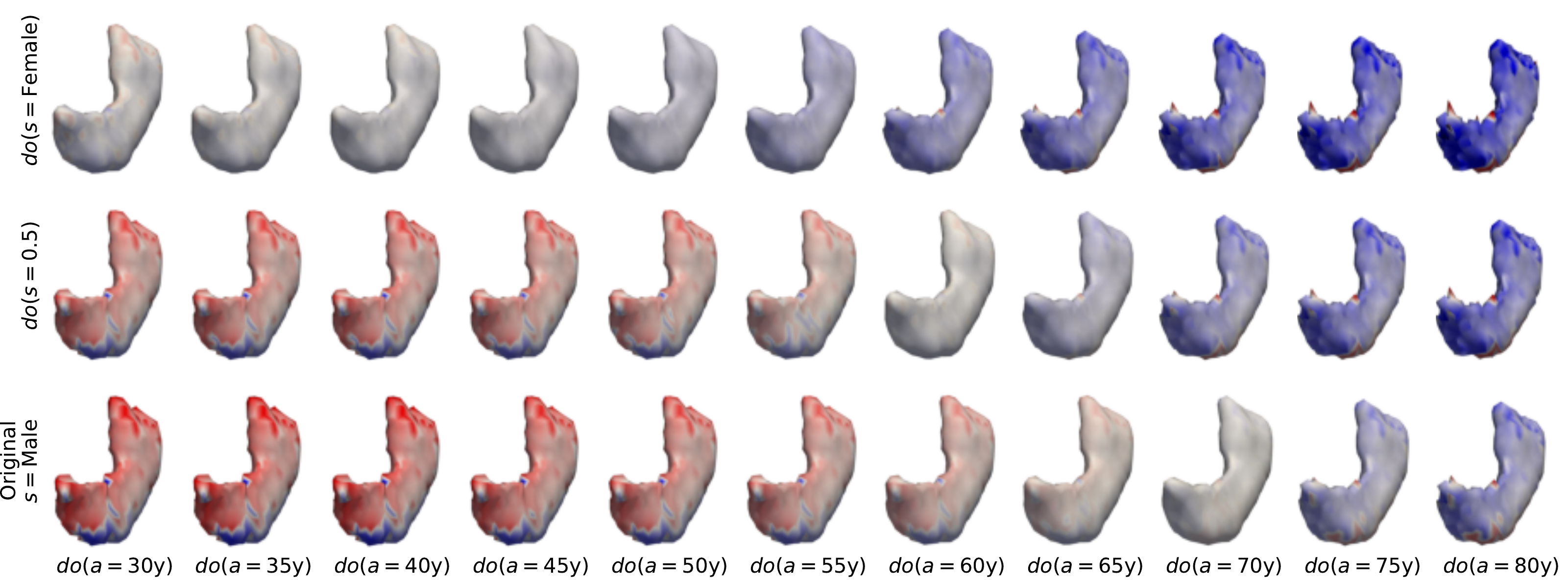}
            }
        \subcaption{Counterfactual meshes for an individual under $do(a)$ and $do(s)$ -- \textit{``What would this person's hippocampus look like if they were older/younger or female?"}. \textbf{Top:} Anterior view. \textbf{Bottom:} Posterior view.}
    \end{subfigure}
    \end{center}
    \begin{center}
        \begin{subfigure}{\textwidth}
            \makebox[\textwidth]{
                \includegraphics[width=0.7\textwidth]{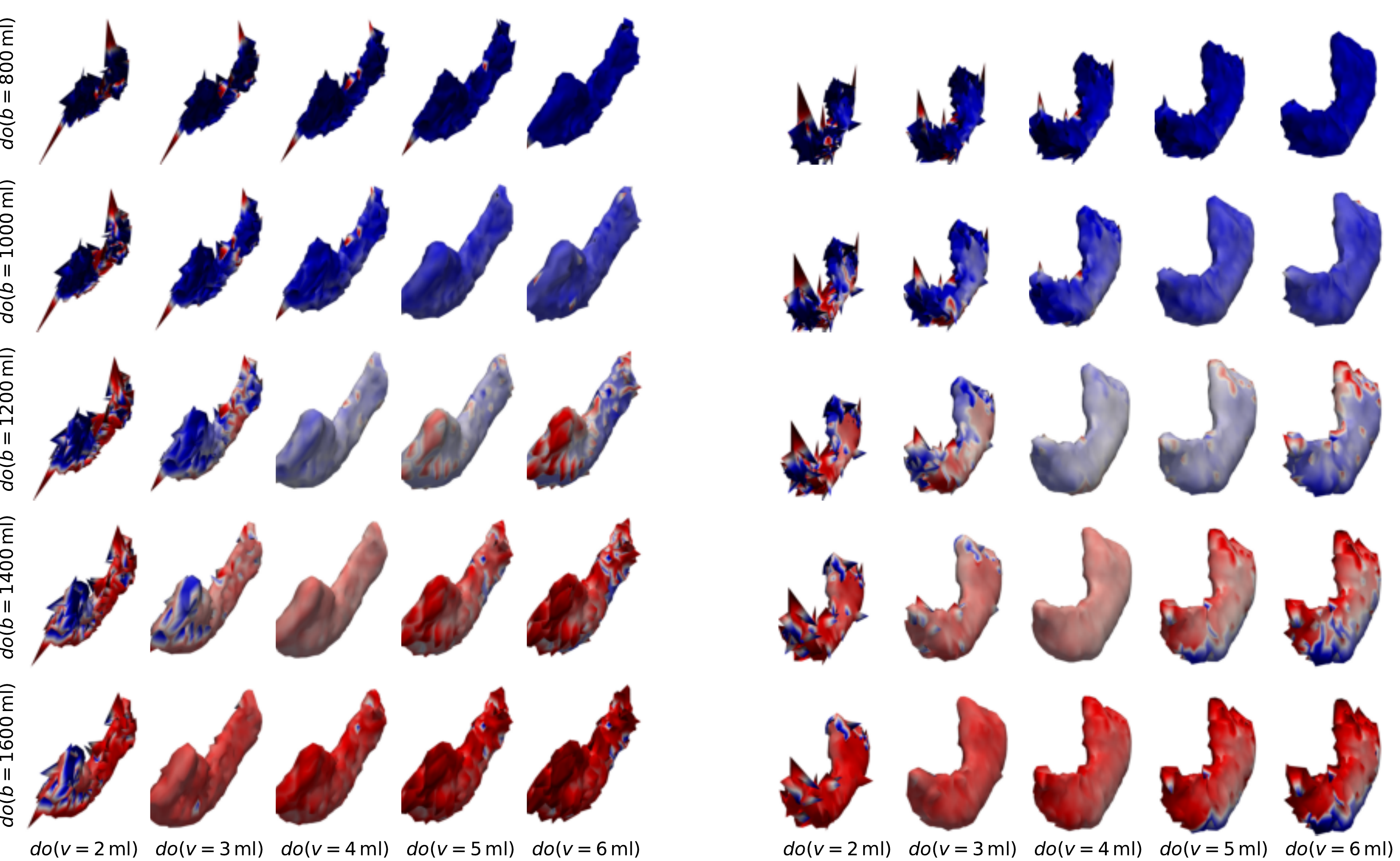}
            }
            \subcaption{Counterfactual meshes for an individual under $do(v)$ and $do(b)$ -- \textit{``What would this person's hippocampus look like if the total brain or hippocampus volumes were larger or smaller?"}.  \textbf{Left:} Anterior view. \textbf{Right:} Posterior view.}
        \end{subfigure}
    \end{center}
\caption{\small Counterfactual hippocampus meshes for a 68 year old, male. Colours show the VED between observed and counterfactual meshes -- \protect\markertwo\, = +5mm to \protect\markerone\, = -5mm.}
\label{fig:hipp_cfs_1}
\end{figure}

\begin{figure}[h]
    \begin{center}
        \begin{subfigure}{\textwidth}
            \makebox[\textwidth]{
                \includegraphics[width=0.7\textwidth]{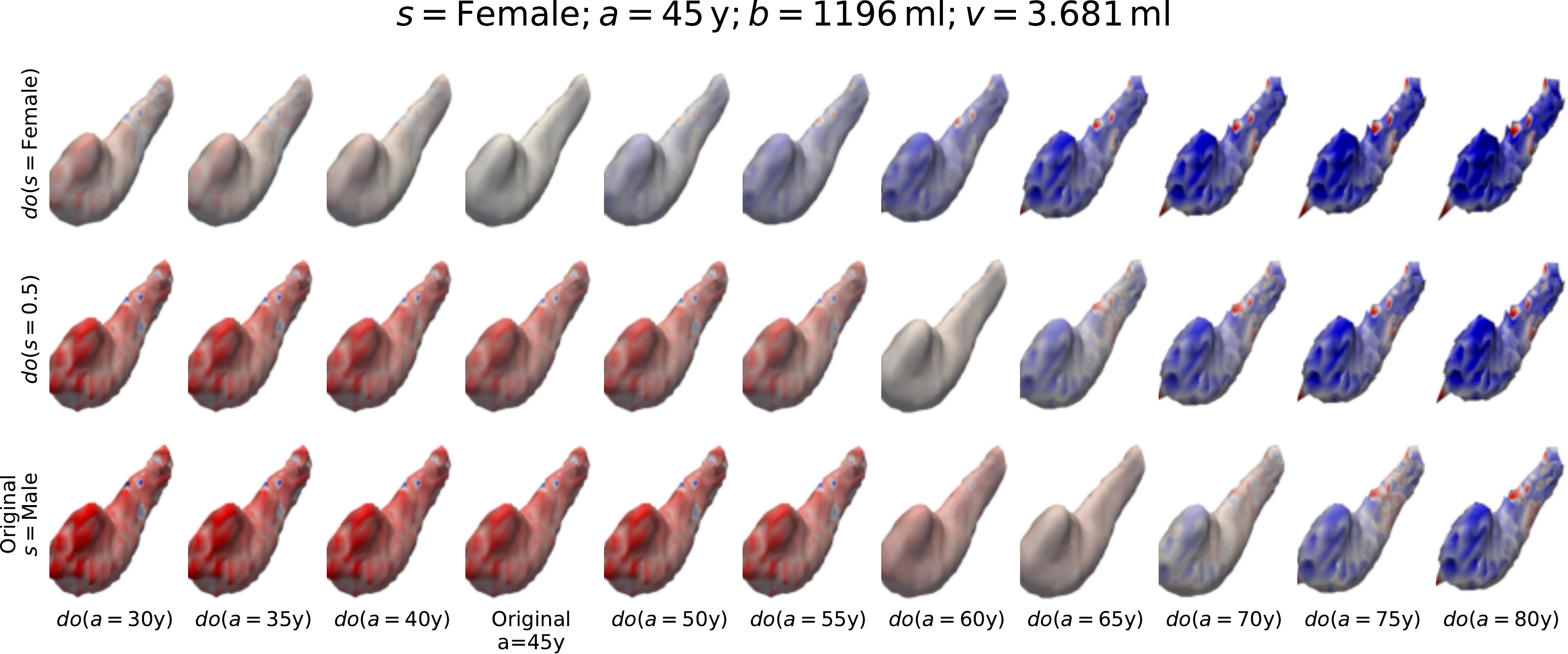}
            }
            \par\bigskip
            \makebox[\textwidth]{
                \includegraphics[width=0.7\textwidth]{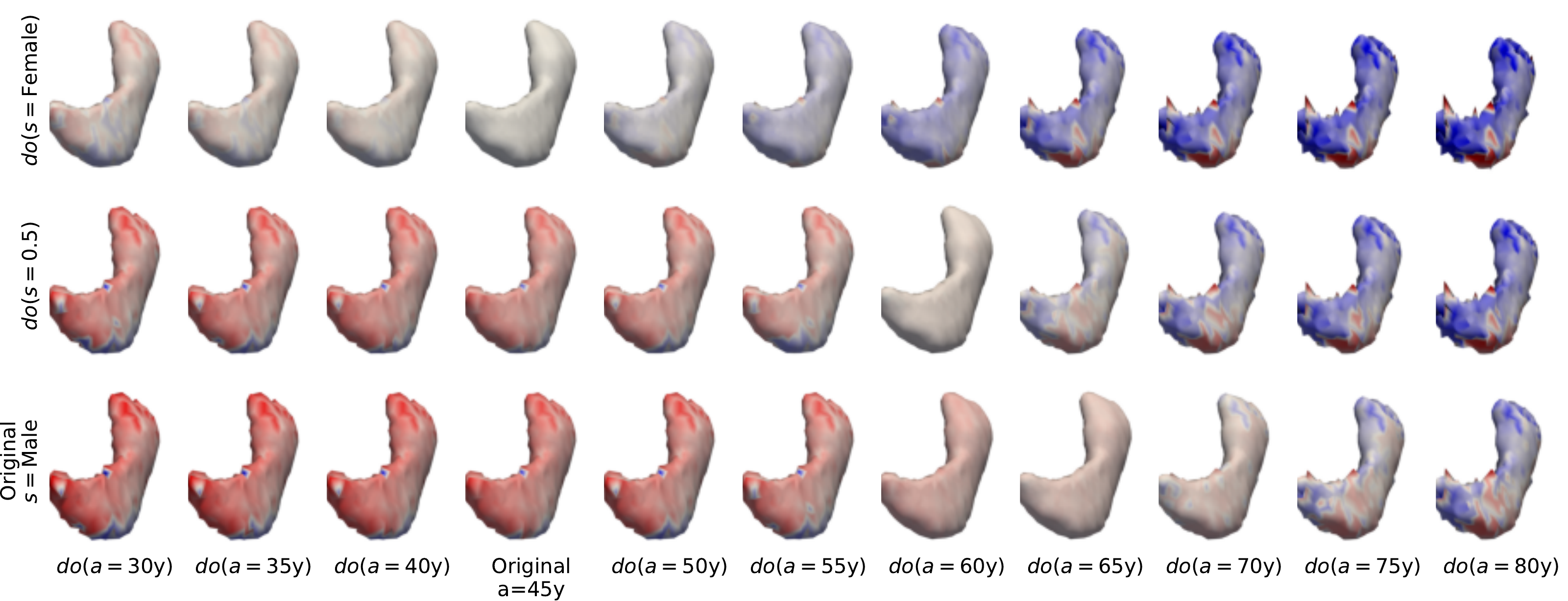}
            }
        \subcaption{Counterfactual meshes for an individual under $do(a)$ and $do(s)$ -- \textit{``What would this person's hippocampus look like if they were older/younger or female?"}. \textbf{Top:} Anterior view. \textbf{Bottom:} Posterior view.}
        \label{fig:medical_age_vvol}
    \end{subfigure}
    \end{center}
    \begin{center}
        \begin{subfigure}{\textwidth}
            \makebox[\textwidth]{
                \includegraphics[width=0.7\textwidth]{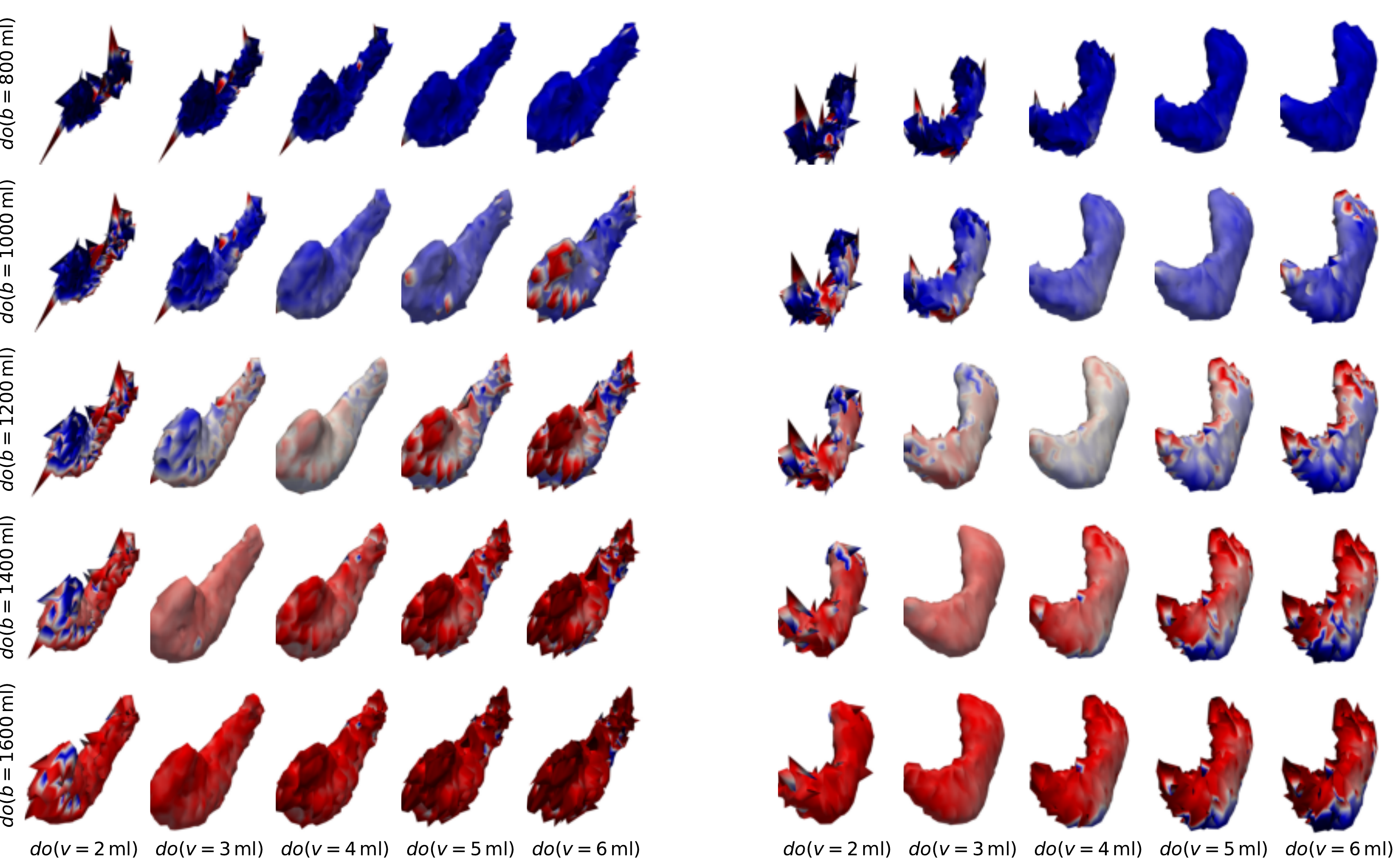}
            }
            \subcaption{Counterfactual meshes for an individual under $do(v)$ and $do(b)$ -- \textit{``What would this person's hippocampus look like if the total brain or hippocampus volumes were larger or smaller?"}. \textbf{Left:} Anterior view. \textbf{Right:} Posterior view.}
            \label{fig:medical_age_bvol}
        \end{subfigure}
    \end{center}
\caption{Counterfactual hippocampus meshes for a 45 year old, female. Colours show the VED between observed and counterfactual meshes -- \protect\markertwo\, = +5mm to \protect\markerone\, = -5mm.}
\label{fig:hipp_cfs_2}
\end{figure}

Hippocampus mesh counterfactuals in \cref{fig:hipp_cfs_1} and \cref{fig:hipp_cfs_2} are produced for the same individuals as \cref{fig:stem_cfs_full_1} and \cref{fig:stem_cfs_full_2}, respectively. Hippocampus meshes are far less regular than the brain stem meshes, resulting in spikes at irregular vertices in counterfactual meshes generated under large interventions, as described in \cref{app:implementation}. However, volume and shape effects remain evidently clear.

\clearpage

